\documentclass[manuscript,screen]{acmart}
\AtBeginDocument{%
  \providecommand\BibTeX{{%
    \normalfont B\kern-0.5em{\scshape i\kern-0.25em b}\kern-0.8em\TeX}}}

\usepackage{booktabs} 

\usepackage[english]{babel}
\usepackage{moresize}
\usepackage{amsmath}
\usepackage{algorithmic}
\usepackage{balance}
\usepackage{comment}
\usepackage{paralist}
\usepackage{bm}
\usepackage{pgfplots}
\usetikzlibrary{pgfplots.dateplot}

\usepackage{fancyhdr}
\usepackage{flushend}
\usepackage[english]{babel}
\usepackage[utf8]{inputenc}
\usepackage{mathrsfs}
\usepackage{graphicx}

\usepackage{amssymb}
\usepackage{amsfonts}
\usepackage{url}
\usepackage{longtable}
\usepackage{rotating}
\usepackage{multirow}
\usepackage{mathrsfs}
\usepackage{subfigure}
\usepackage{enumitem}
\usepackage[linesnumbered,algoruled,boxed,lined]{algorithm2e}
\usepackage{adjustbox}
\usepackage{hyperref}
\usepackage{pgfplots}
\usetikzlibrary{pgfplots.dateplot}
\usepackage{filecontents}
\definecolor{tblue}{RGB}{31,119,180}
\definecolor{torange}{RGB}{255,127,14}
\definecolor{tgreen}{RGB}{44,160,44}
\definecolor{tred}{RGB}{214,39,40}
\definecolor{tpurple}{RGB}{148,103,189}

\usepackage{pifont}
\usepackage{makecell}
\usepackage{wrapfig}
\usepackage{colortbl}
\definecolor{mygray}{gray}{.85}
\definecolor{mygray1}{gray}{.7}
\definecolor{mygray2}{gray}{.93}

\usepackage{listings}
\usepackage{etoolbox}
\makeatletter
\AfterEndEnvironment{algorithm}{\let\@algcomment\relax}
\AtEndEnvironment{algorithm}{\kern2pt\hrule\relax\vskip3pt\@algcomment}
\let\@algcomment\relax
\newcommand\algcomment[1]{\def\@algcomment{\footnotesize#1}}
\renewcommand\fs@ruled{\def\@fs@cfont{\bfseries}\let\@fs@capt\floatc@ruled
  \def\@fs@pre{\hrule height.8pt depth0pt \kern2pt}%
  \def\@fs@post{}%
  \def\@fs@mid{\kern2pt\hrule\kern2pt}%
  \let\@fs@iftopcapt\iftrue}
\makeatother

\usepackage{tabulary}
\newcolumntype{I}{!{\vrule width 1pt}}
\newcolumntype{x}[1]{>{\centering\arraybackslash}p{#1pt}}
\newcolumntype{y}[1]{>{\raggedright\arraybackslash}p{#1pt}}
\newcolumntype{z}[1]{>{\raggedleft\arraybackslash}p{#1pt}}

\definecolor{codegreen}{RGB}{79,126,127}
\definecolor{codedefine}{RGB}{153,54,159}
\definecolor{codefunc}{RGB}{73,122,234}
\definecolor{codecall}{RGB}{73,122,234}
\definecolor{codepro}{RGB}{212,96,80}
\definecolor{codedim}{RGB}{89,152,195}
\definecolor{3dgc1}{RGB}{177, 83, 74}
\definecolor{3dgc2}{RGB}{93, 107, 72}

\newcommand{\pub}[1]{{\color{gray}{\tiny{[{#1}]\!}}}}

\makeatletter
\newcommand{\thickhline}{%
    \noalign {\ifnum 0=`}\fi \hrule height 1pt
    \futurelet \reserved@a \@xhline
}
\makeatother

\makeatletter
\DeclareRobustCommand\onedot{\futurelet\@let@token\@onedot}
\def\@onedot{\ifx\@let@token.\else.\null\fi\xspace}

\def\eg{\emph{e.g}\onedot} 
\def\ie{\emph{i.e}\onedot} 
\def\cf{\emph{c.f}\onedot} 
\def\etc{\emph{etc}\onedot} \def\vs{\emph{vs}\onedot}
 
\def\etal{\emph{et al}\onedot}

\makeatother

\setcopyright{acmcopyright}

\begin{document}

\title{A Survey on 3D Gaussian Splatting}

\author{Guikun Chen}
\affiliation{%
  \institution{The State Key Lab of Brain-Machine Intelligence, Zhejiang University}
  \city{Hangzhou}
  \country{China}}
\email{guikunchen@gmail.com}

\author{Wenguan Wang}
\authornote{Corresponding author}
\affiliation{%
\institution{The State Key Lab of Brain-Machine Intelligence, Zhejiang University}
\city{Hangzhou}
\country{China}}
\email{wenguanwang.ai@gmail.com}


\begin{abstract}
3D Gaussian splatting (GS) has emerged as a transformative technique in radiance fields. Unlike mainstream implicit neural models, 3D GS uses millions of learnable 3D Gaussians for an explicit scene representation. Paired with a differentiable rendering algorithm, this approach achieves real-time rendering and unprecedented editability, making it a potential game-changer for 3D reconstruction and representation.
In the present paper, we provide the first systematic overview of the recent developments and critical contributions in 3D GS. We begin with a detailed exploration of the underlying principles and the driving forces behind the emergence of 3D GS, laying the groundwork for understanding its significance.
A focal point of our discussion is the practical applicability of 3D GS. By enabling unprecedented rendering speed, 3D GS opens up a plethora of applications, ranging from virtual reality to interactive media and beyond. This is complemented by a comparative analysis of leading 3D GS models, evaluated across various benchmark tasks to highlight their performance and practical utility.
The survey concludes by identifying current challenges and suggesting potential avenues for future research. Through this survey, we aim to provide a valuable resource for both newcomers and seasoned researchers, fostering further exploration and advancement in explicit radiance field.
\end{abstract}




\maketitle

\section{Introduction}
\label{sec:intro}

The objective of image based 3D scene reconstruction is to convert a collection of views capturing a scene into a digital 3D model that can be computationally processed, analyzed, and manipulated. This hard and long-standing problem is fundamental for machines to comprehend real-world environments, facilitating a wide array of applications such as 3D modeling and animation, robot navigation, historical preservation, augmented/virtual reality, and autonomous driving.

The journey of 3D scene reconstruction began long before the surge of deep learning, with early endeavors focusing on light fields and basic reconstruction methods~\cite{gortler2023lumigraph,levoy2023light,buehler2023unstructured}. These early attempts, however, were limited by their reliance on dense sampling and structured capture, leading to challenges in handling complex scenes and lighting conditions. The emergence of structure-from-motion~\cite{snavely2006photo} and subsequent advancements in multi-view stereo~\cite{goesele2007multi} algorithms provided a more robust framework. Despite these advancements, such methods struggled with novel-view synthesis and texture loss. NeRF represents a quantum leap in this progression. By leveraging neural networks, NeRF enabled the direct mapping of spatial coordinates to color and density. The success of NeRF hinged on its ability to create continuous, volumetric scene functions, producing results with unprecedented fidelity. However, as with any burgeoning technology, it came at a cost: \textbf{i}) Computational Intensity. NeRF based methods are computationally intensive~\cite{garbin2021fastnerf,muller2022instant}, often requiring substantial training times and resources for rendering, especially for high-resolution outputs. \textbf{ii}) Editability. Manipulating scenes represented implicitly is challenging, since direct modifications to the neural network's weights are not intuitively related to changes in geometric or appearance properties of the scene.

It is in this context that 3D Gaussian splatting (GS)~\cite{kerbl20233d} emerges, not merely as an incremental improvement but as a paradigm-shifting approach that redefines the boundaries of scene representation and rendering. While NeRF excelled in creating photorealistic images, the need for faster, more efficient rendering methods was becoming increasingly apparent, especially for applications (\eg, virtual reality and autonomous driving) that are highly sensitive to latency. 3D GS addressed this need by introducing an advanced, explicit scene representation that models a scene using millions of learnable 3D Gaussians in space. Unlike the implicit, coordinate-based models~\cite{sitzmann2019deepvoxels,mildenhall2020nerf}, 3D GS employs an explicit representation and highly parallelized workflows, facilitating more efficient computation and rendering. The innovation of 3D GS lies in its unique blend of the benefits of differentiable pipelines and point-based rendering techniques~\cite{pfister2000surfels,zwicker2001surface,ren2002object,yifan2019differentiable}. By representing scenes with learnable 3D Gaussians, it preserves the strong fitting capability of continuous volumetric radiance fields, essential for high-quality image synthesis, while simultaneously avoiding the computational overhead associated with NeRF based methods (\eg, computationally expensive ray-marching, and unnecessary calculations in empty space).

The introduction of 3D GS is not just a technical advancement; it represents a fundamental shift in how we approach scene representation and rendering in computer vision and graphics. By enabling real-time rendering capabilities without compromising on visual quality, 3D GS opens up a plethora of possibilities for applications ranging from virtual reality and augmented reality to real-time cinematic rendering and beyond~\cite{kalkofen2008comprehensible,patney2016towards,albert2017latency,jiang2024vr}. This technology holds the promise of not only enhancing existing applications but also enabling new ones that were previously unfeasible due to computational constraints. Furthermore, 3D GS's explicit scene representation offers unprecedented flexibility to control the objects and scene dynamics, a crucial factor in complex scenarios involving intricate geometries and varying lighting conditions~\cite{lu2023scaffold,saito2023relightable,zhang2024darkgs}. This level of editability, combined with the efficiency of the training and rendering process, positions 3D GS as a transformative force in shaping future developments in relevant fields.

In an effort to assist readers in keeping pace with the swift evolution of 3D GS, we provide the first survey on 3D GS, which presents a systematic and timely collection of the most significant literature on the topic. Given that GS is a very recent innovation (\cf Fig.~\ref{fig_num_paper}), this survey focuses in particular on its principles, and the diverse developments and contributions that have emerged since its introduction. We systematically review the foundational period of GS, with selected works primarily sourced from top-tier conferences. Our analysis concentrates on the initial explosion of research following its introduction in 2023 through 2024, covering the theoretical underpinnings, landmark developments, and early applications that shaped the field. Acknowledging the nascent yet rapidly evolving nature, this survey is inevitably a biased view, but we strive to offer a balanced perspective that reflects both the current state and the future potential of this field. Our aim is to encapsulate the primary research trends and serve as a valuable resource for both researchers and practitioners eager to understand and contribute to this rapidly evolving domain. The distinctions of this paper from existing surveys (general~\cite{fei20243d,dalal2024gaussian,bao20243d,wu2024recent} or domain-specific~\cite{bagdasarian20253dgs,ali2025compression,he2025survey}) are evident in the following aspects:

\noindent$\bullet$ We provide the first systematic and comprehensive review that examines 3D GS from a macro-level perspective by establishing clear taxonomies and frameworks. This high-level systematization helps researchers identify trends and potential directions that might not be apparent from paper-specific reviews. Our organizational structure serves as a roadmap for understanding how different approaches relate to and build upon each other within the 3D GS ecosystem.

\noindent$\bullet$ This paper is the first and only survey to thoroughly delve into the theoretical background and fundamental principles of 3D GS. The comprehensive coverage makes the field more approachable for newcomers while providing valuable insights for experienced researchers.

\noindent$\bullet$ To ensure our survey remains relevant and offer long-term value in this rapidly evolving field, we maintain two dynamic GitHub repositories: \href{https://github.com/guikunchen/Awesome3DGS}{one} that follows our survey's organizational structure and \href{https://github.com/guikunchen/3DGS-Benchmarks}{another} that includes comprehensive performance comparisons with analysis data.

\begin{figure}[t]
\centering
\includegraphics[width=1.0\textwidth]{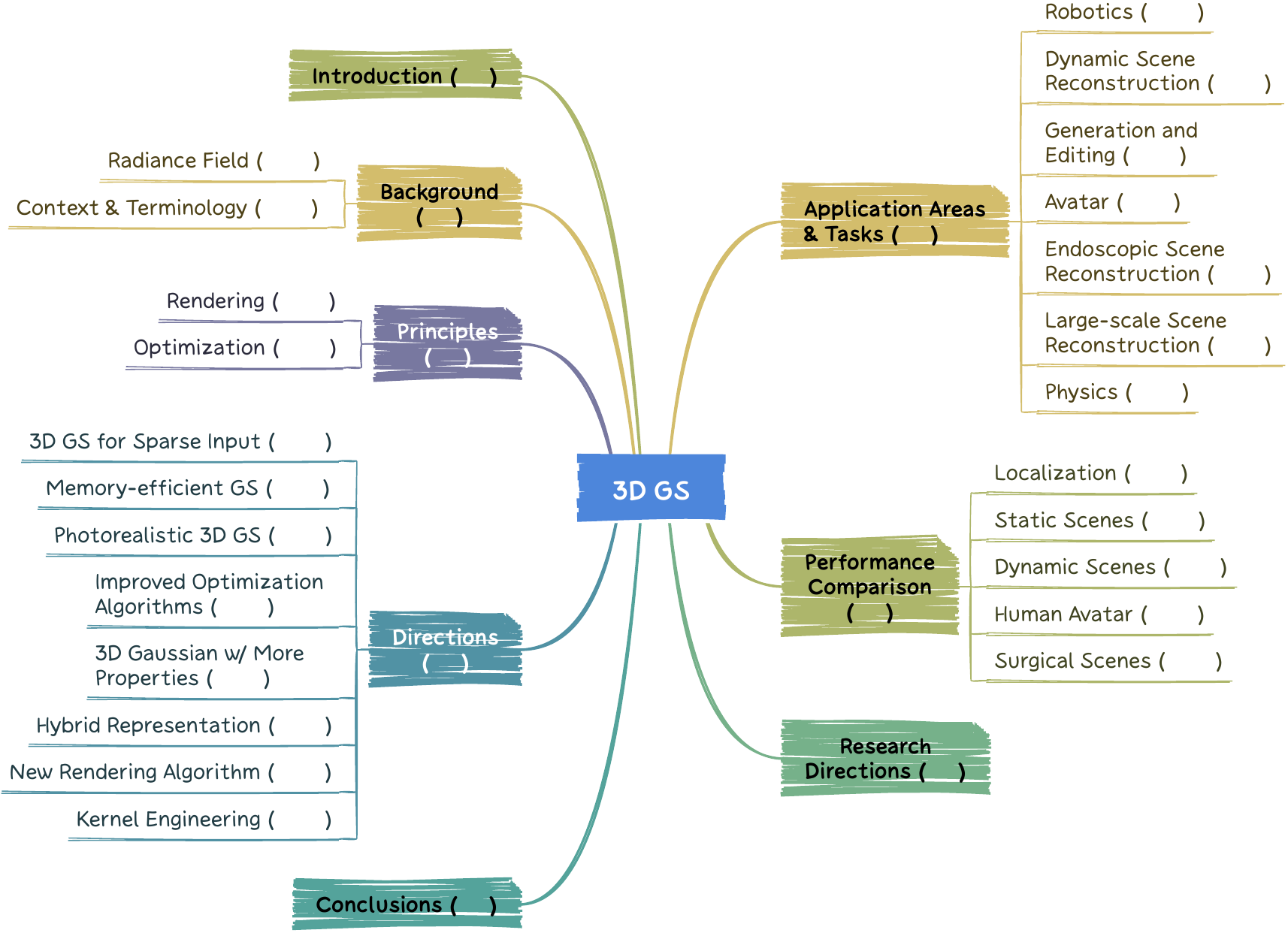}
\put(-275.7, 283.0){{\S\ref{sec:intro}}}
\put(-287.2, 235.9){{\S\ref{sec:background}}}
\put(-284.8, 189.6){\textcolor{white}{\S\ref{sec:3dgs}}}
\put(-285.5, 86.9){\textcolor{white}{\S\ref{sec:e3dgs}}}
\put(-128.7, 230.1){{{\S\ref{sec:application}}}}
\put(-143.6, 104.1){{{\S\ref{sec:performance}}}}
\put(-119.7, 50.7){{\S\ref{sec:direction}}}
\put(-276.1, 6){{\S\ref{sec:conclusions}}}
\put(-340.6, 254.7){{\S\ref{sec:bg_radiance_field}}}
\put(-341.5, 238.8){{\S\ref{sec:background_context}}}
\put(-335.7, 208.1){{\S\ref{sec:3dgs_nvs}}}
\put(-335.5, 192.2){{\S\ref{sec:3dgs_optm}}}
\put(-337, 161.0){{\S\ref{sec:e3dgs_de3dgs}}}
\put(-337.8, 145.7){{\S\ref{sec:e3dgs_me3dgs}}}
\put(-336.8, 130.0){{\S\ref{sec:e3dgs_photo}}}
\put(-356.5, 106.0){{\S\ref{sec:e3dgs_optim}}}
\put(-357.5, 82.3){{\S\ref{sec:e3dgs_property}}}
\put(-336.9, 66.2){{\S\ref{sec:e3dgs_struct}}}
\put(-336.9, 50.8){{\S\ref{sec:e3dgs_new_rendering}}}
\put(-337, 35.0){{\S\ref{sec:kernel_engineering}}}
\put(-46.2, 304.7){{\S\ref{sec:app_robotics}}}
\put(-23.7, 280.7){{\S\ref{sec:app_dynamic_scene}}}
\put(-52.0, 256.6){{\S\ref{sec:app_gen_edit}}}
\put(-54.2, 241.1){{\S\ref{sec:app_avatar}}}
\put(-23.5, 216.9){{\S\ref{sec:app_endo}}}
\put(-23.5, 193.8){{\S\ref{sec:app_largescale}}}
\put(-51.4, 177.7){{\S\ref{sec:app_physics}}}
\put(-51.2, 150.6){{\S\ref{sec:perm_local}}}
\put(-45.5, 134.7){{\S\ref{sec:perm_render_static}}}
\put(-38.4, 119.3){{\S\ref{sec:perm_render_dynamic}}}
\put(-45.6, 102.9){{\S\ref{sec:perm_avatar}}}
\put(-40.0, 87.6){{\S\ref{sec:perm_surgical}}}

\vspace{-1em}
\caption{Structure of the overall review.}
\vspace{-1em}
\label{fig_structure}
\end{figure}

A summary of the structure of this article can be found in Fig.~\ref{fig_structure}, which is presented as follows: Sec.~\ref{sec:background} provides a brief background on problem formulation, terminology, and related research domains. Sec.~\ref{sec:3dgs} introduces the essential insights of 3D GS, encompassing the rendering process with learned 3D Gaussians and the optimization details (\ie, how to learn 3D Gaussians) of 3D GS. Sec.~\ref{sec:e3dgs} presents several fruitful directions that aim to improve the capabilities of the original 3D GS. Sec.~\ref{sec:application} unveils the diverse application areas and tasks where 3D GS has made significant impacts, showcasing its versatility. Sec.~\ref{sec:performance} conducts performance comparison and analysis. Finally, Sec.~\ref{sec:direction} and~\ref{sec:conclusions} highlight the open questions for further research and conclude the survey.

\section{Background}
\label{sec:background}

This section provides a brief formulation of radiance fields (Sec. \ref{sec:bg_radiance_field}), including both implicit and explicit ones. Sec.~\ref{sec:background_context} further establishes linkages with relevant rendering algorithms and terminologies. Please see~\cite{kobbelt2004survey,xie2022neural,wang2024visual,tewari2022advances,han2019image} for a comprehensive overview of radiance fields, scene reconstruction and representation, and rendering methods.

\subsection{Radiance Field}
\label{sec:bg_radiance_field}

\noindent$\bullet$~\textbf{Implicit Radiance Field.}
An implicit radiance field represents light distribution in a scene without explicitly defining the geometry of the scene. In the deep learning era, neural networks are often used to learn a continuous volumetric scene representation~\cite{mescheder2019occupancy,park2019deepsdf}. The most prominent example is NeRF~\cite{mildenhall2020nerf}. In NeRF (Fig.~\ref{fig:fig_nerf_3dgs}a), one or more MLPs are used to map a set of spatial coordinates \( (x,y,z) \) and viewing directions \( (\theta, \phi) \) to color $c$ and volume density $\sigma$:
\begin{equation}
    (c, \sigma) \leftarrow \text{MLP}(x, y, z, \theta, \phi).
\end{equation}
This format allows for a differentiable and compact representation of scenes, albeit often at the cost of high computational load. Note that typically, the color $c$ is direction-dependent, whereas the volume density $\sigma$ is not~\cite{mildenhall2020nerf}.

\noindent$\bullet$~\textbf{Explicit Radiance Field.}
An explicit radiance field directly represents the distribution of light in a discrete spatial structure, such as a voxel grid or a set of points~\cite{sun2022direct,fridovich2023k}. Each element in this structure stores the radiance information for its respective location. This allows for direct and often faster access to radiance data but at the cost of higher memory usage and potentially lower resolution. Plenoxels~\cite{fridovich2022plenoxels} highlighted this trend by discarding the MLP and directly optimizing a sparse voxel grid with spherical harmonic coefficients, showing that a fully explicit grid can achieve NeRF-level quality while converging much faster. Similar to the implicit radiance field, the explicit one is written as:
\begin{equation}
    (c, \sigma) \leftarrow \text{DataStructure}(x, y, z, \theta, \phi),
\end{equation}
where $\text{DataStructure}$ could be in the format of volumes, point clouds, \etc $\text{DataStructure}$ encodes directional color in two main ways. One is encoding high-dimensional features that are subsequently decoded by a lightweight MLP. Another one is directly storing coefficients of directional basis functions, such as spherical harmonics or spherical Gaussians, where the final color is computed as a function of these coefficients and the viewing direction.

\noindent$\bullet$~\textbf{3D Gaussian Splatting: Best-of-Both Worlds.}
3D GS~\cite{kerbl20233d} is an explicit radiance field with the advantages of implicit radiance fields. Concretely, it uses the strengths of both paradigms by utilizing \textit{learnable} 3D Gaussians as the basis elements of $\text{DataStructure}$. Note that 3D GS encodes the opacity $\alpha$ directly for each Gaussian, as opposed to approaches of first establishing density $\sigma$ and then computing opacity based on that density. As in previous reconstruction work, 3D Gaussians are optimized under the supervision of multi-view images to represent the scene. Such a 3D Gaussian based differentiable pipeline combines the benefits of neural network based optimization and explicit, structured data storage. This hybrid approach aims to achieve real-time, high-quality rendering and requires less training time, particularly for complex scenes and high-resolution outputs. In addition, from a differentiable rendering perspective, 3D GS complements mesh-based differentiable rasterizers such as nvdiffrast~\cite{laine2020modular}, which operate on explicit triangle meshes using high-performance modular rasterization primitives.

\subsection{Context and Terminology}
\label{sec:background_context}

\noindent$\bullet$~\textbf{Volumetric rendering}
aims to transform a 3D volumetric representation into an image by integrating radiance along camera rays. A camera ray $\bm{r}(t)$ can be parameterized as: $\bm{r}(t) \!=\! \bm{o} \!+\! t\bm{d}, t \!\in\! [t_{\text{near}}, t_{\text{far}}],$
where $\bm{o}$ represents the ray origin (camera center), $\bm{d}$ is the ray direction, and $t$ indicates the distance along the ray between near and far clipping planes. The pixel color $C(\bm{r})$ is computed through a line integral along the ray $\bm{r}(t)$, mathematically expressed as~\cite{mildenhall2020nerf}:
\begin{equation}
\label{eq:volume_rendering}
    C(\bm{r}) = \int_{t_{\text{near}}}^{t_{\text{far}}} T(t) \, \sigma(\bm{r}(t)) \, c(\bm{r}(t), \bm{d}) \, dt,
\end{equation}
where $\sigma(\bm{r}(t))$ is the volume density at point $\bm{r}(t)$, $c(\bm{r}(t), \bm{d})$ is the color at that point, and $T(t)$ is the transmittance. Ray-marching directly approximates the volumetric rendering integral by systematically ``stepping'' along a ray and sampling the scene's properties at discrete intervals. NeRF~\cite{mildenhall2020nerf} shares the same spirit of ray-marching and introduces importance sampling and positional encoding to improve the quality of synthesized images. While providing high-quality results, ray-marching is computationally expensive, especially for high-resolution images.

\noindent$\bullet$~\textbf{Point-based rendering}
represents another class of rendering algorithms, of which 3D GS introduces a notable implementation. Its simplest form~\cite{grossman1998point} rasterizes point clouds with a fixed size, which introduces drawbacks such as holes and rendering artifacts. Seminal works addressed these limitations through various methods, including: \textbf{i}) employing multi-resolution level-of-detail hierarchies like QSplat~\cite{rusinkiewicz2000qsplat} to handle large-scale scenes, \textbf{ii}) splatting point primitives with a spatial extent~\cite{zwicker2001surface,zwicker2001ewa,zwicker2002ewa,ren2002object}, and \textbf{iii}) more recently, embedding neural features directly into points for subsequent network-based rendering~\cite{aliev2020neural,ruckert2022adop}. 3D GS uses 3D Gaussian as the point primitive that contains explicit attributes (\eg, color and opacity) instead of implicit neural features. The rendering approach, \ie, point-based $\alpha$-blending (exemplified in Eq.~\ref{eq:alpha_compo}), shares the same image formation model as NeRF-style volumetric rendering (Eq.~\ref{eq:volume_rendering})~\cite{kerbl20233d}, but demonstrates substantial speed advantages. This advantage originates from fundamental algorithmic differences. NeRFs approximate a line integral along a ray for each pixel, requiring expensive sampling. Point-based methods render point clouds using rasterization, which inherently benefits from parallel computational strategies~\cite{lassner2021pulsar}.

\section{3D Gaussian Splatting: Principles}
\label{sec:3dgs}

3D GS offers a breakthrough in real-time, high-resolution image rendering, without relying on deep neural networks. This section aims to provide essential insights of 3D GS. We first elaborate on how 3D GS synthesizes an image given well-constructed 3D Gaussians in Sec.~\ref{sec:3dgs_nvs}, \ie, the forward process of 3D GS. Then, we introduce how to obtain well-constructed 3D Gaussians for a given scene in Sec.~\ref{sec:3dgs_optm}, \ie, the optimization process of 3D GS.

\subsection{Rendering with Learned 3D Gaussians}
\label{sec:3dgs_nvs}

Consider a scene represented by optimized 3D Gaussians. The objective is to generate an image from a specified camera pose. Recall that NeRFs approach this task through volumetric ray-marching, 
sampling 3D space points per pixel. Such a paradigm struggles with high-resolution image synthesis, failing to achieve real-time rendering, especially for platforms with limited computing resources~\cite{kerbl20233d}. By contrast, 3D GS begins by projecting these 3D Gaussians onto a pixel-based image plane, a process termed ``splatting''~\cite{zwicker2001ewa,zwicker2002ewa} (\cf Fig.~\ref{fig:fig_nerf_3dgs}b). Afterwards, 3D GS sorts these Gaussians and computes the value for each pixel. As shown in Fig.~\ref{fig:fig_nerf_3dgs}, the rendering of NeRFs and 3D GS can be viewed as an inverse process of each other. In what follows, we begin with the definition of a 3D Gaussian, which is the minimal element of the scene representation in 3D GS. Next, we describe how these 3D Gaussians can be used for differentiable rendering. Finally, we introduce the acceleration technique used in 3D GS, which is the key to fast rendering.

\begin{figure*}[t]
\centering
\includegraphics[width=0.99\textwidth]{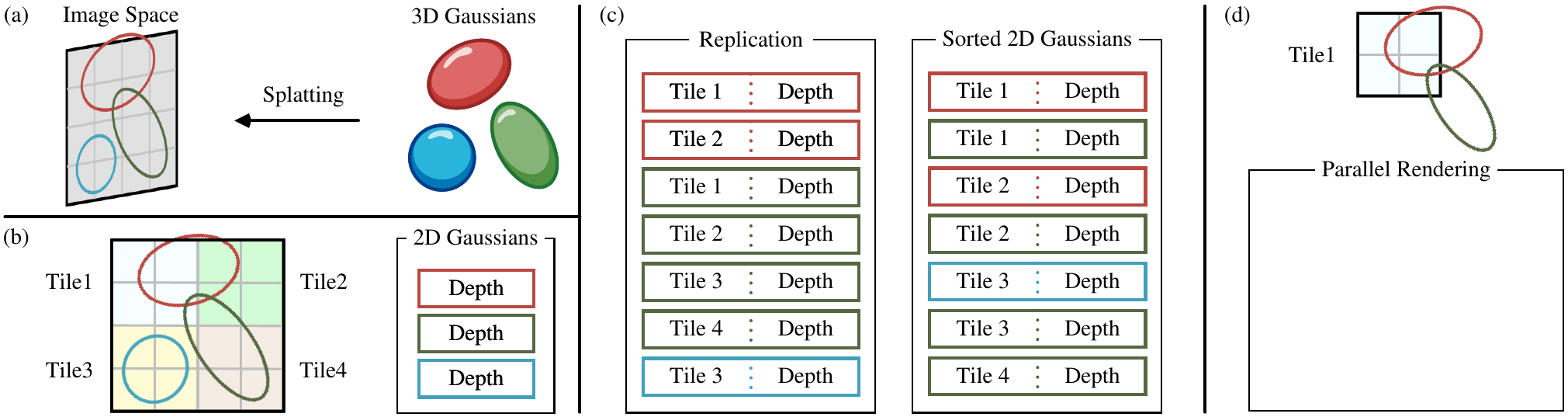}
\put(-55, 102){\footnotesize\textcolor{black}{$C_1 ~~ C_2$}}
\put(-55, 91){\footnotesize\textcolor{black}{$C_3 ~~ C_4$}}
\put(-81.5, 50){\footnotesize{$C_1 \!=\! \textcolor{3dgc1}{\alpha_1'} c_1 + \textcolor{3dgc2}{\alpha_1'} c_2 (1 - \textcolor{3dgc1}{\alpha_1'})$}}
\put(-81.5, 36){\footnotesize{$C_2 \!=\! \textcolor{3dgc1}{\alpha_2'} c_1 + \textcolor{3dgc2}{\alpha_2'} c_2 (1 - \textcolor{3dgc1}{\alpha_2'})$}}
\put(-81.5, 22){\footnotesize{$C_3 \!=\! \textcolor{3dgc1}{\alpha_3'} c_1 + \textcolor{3dgc2}{\alpha_3'} c_2 (1 - \textcolor{3dgc1}{\alpha_3'})$}}
\put(-81.5, 8){\footnotesize{$C_4 \!=\! \textcolor{3dgc1}{\alpha_4'} c_1 + \textcolor{3dgc2}{\alpha_4'} c_2 (1 - \textcolor{3dgc1}{\alpha_4'})$}}

\vspace{-8pt}
\caption{An illustration of the forward process of 3D GS (see Sec.~\ref{sec:3dgs_nvs}). (a) The splatting step projects 3D Gaussians into image space. (b) 3D GS divides the image into multiple non-overlapping patches, \ie, tiles. (c) 3D GS replicates the Gaussians which cover several tiles, assigning each copy an identifier, \ie, a tile ID. (d) By rendering the sorted Gaussians, we can obtain all pixels within the tile. Note that the computational workflows for pixels and tiles are independent and can be done in parallel. Best viewed in color.}
\label{fig:fig_3dgs}
\vspace{-1.5em}
\end{figure*}

\noindent$\bullet$~\textbf{Properties of 3D Gaussian.} A 3D Gaussian is characterized by its center $\bm{\mu}$, opacity $\alpha$, 3D covariance matrix $\bm{\Sigma}$, and color $c$. $c$ is represented by spherical harmonics (SH) coefficients to model view-dependent appearance. Unlike a static RGB value, SH functions allow the color to vary based on the viewing angle. This enables the representation to capture non-Lambertian effects, such as specular highlights and lighting variations, by evaluating the SH coefficients with the specific view direction during rendering. All the properties are learnable and optimized through back-propagation.

\noindent$\bullet$~\textbf{Frustum Culling.} Given a specified camera pose, this step determines which 3D Gaussians are outside the camera's frustum. By doing so, 3D Gaussians outside the given view will not be involved in the subsequent computation.

\noindent$\bullet$~\textbf{Splatting} (Fig.~\ref{fig:fig_3dgs}a). In this step, 3D Gaussians (ellipsoids) in 3D space are projected into 2D image space (ellipses). The projection proceeds through two transformations: first, transforming 3D Gaussians from world coordinates to camera coordinates using viewing transformation, and subsequently splatting these Gaussians into 2D image space via an approximation of projective transformation. Given the 3D covariance matrix $\bm{\Sigma}$ describing a 3D Gaussian and viewing transformation matrix $\bm{W}$, the 2D covariance matrix $\bm{\Sigma}'$ characterizing the projected 2D Gaussian is computed through:
\begin{equation}
    \bm{\Sigma}' = \bm{JW\Sigma}\bm{W}^{\top}\bm{J}^{\top},
\end{equation}
where $\bm{J}$ is the Jacobian of the affine approximation of the projective transformation~\cite{zwicker2001ewa,kerbl20233d}. One might wonder why the standard camera intrinsics based projective transformation is not used here. This is because its mappings are not affine and therefore cannot directly project $\bm{\Sigma}$. 3D GS adopts an affine one proposed in~\cite{zwicker2001ewa} which approximates the projective transformation using the first two terms (including $\bm{J}$) of the Taylor expansion (see Sec. 4.4 in~\cite{zwicker2001ewa}).

\noindent$\bullet$~\textbf{Rendering by Pixels.} Before delving into the final version of GS with highly parallel computation, we first elaborate on its simpler form to offer insights into its basic working mechanism. Given the position of a pixel $\bm{x}$, its distance to all overlapping Gaussians, \ie, the depths of these Gaussians, can be computed through the viewing transformation matrix $\bm{W}$, forming a sorted list of Gaussians $\mathcal{N}$. Then, $\alpha$-blending~\cite{porter1984compositing} is adopted to compute the final color of this pixel:
\begin{equation}
\label{eq:alpha_compo}
    C = \sum\nolimits_{n = 1 }^{|\mathcal{N}|} c_n \alpha'_n \prod\nolimits_{j=1}^{n-1}\left(1-\alpha'_j\right),
\end{equation}
where $c_n$ is the learned color. The final opacity $\alpha'_n$ is the multiplication result of the learned opacity $\alpha_n$ and the Gaussian:
\begin{equation}
\label{eq:gaussian_opacity}
    \alpha'_n = \alpha_n \times \text{exp}\big(-0.5(\bm{x}' - \bm{\mu}'_n)^{\top}\bm{\Sigma}'^{-1}_{n}(\bm{x}' - \bm{\mu}'_n)\big),
\end{equation}
where $\bm{x}'$ and $\bm{\mu}'_n$ are coordinates in the projected space. It is a reasonable concern that the rendering process described could be slower compared to NeRFs, given that generating the required sorted list is hard to parallelize. Indeed, this concern is justified; rendering speeds can be significantly impacted when utilizing such a simplistic, pixel-by-pixel approach. To achieve real-time rendering, 3D GS makes several concessions to accommodate parallel computation.

\noindent$\bullet$~\textbf{Tiles (Patches).}
To avoid the cost computation of deriving Gaussians for each pixel, 3D GS shifts the precision from pixel-level to patch-level detail, which is inspired by tile-based rasterization~\cite{lassner2021pulsar}. Concretely, 3D GS initially divides the image into multiple non-overlapping patches (tiles). Fig.~\ref{fig:fig_3dgs}b provides an illustration of tiles. Each tile comprises 16$\times$16 pixels as suggested in~\cite{kerbl20233d}. 3D GS further determines which \textbf{tiles} intersect with these projected Gaussians. Given that a projected Gaussian may cover several tiles, a logical method involves replicating the Gaussian, assigning each copy an identifier (\ie, a tile ID) for the relevant tile.

\noindent$\bullet$~\textbf{Parallel Rendering.}
After replication, 3D GS combines the respective tile ID with the depth value obtained from the view transformation for each Gaussian. This results in an unsorted list of bytes where the upper bits represent the tile ID and the lower bits signify depth. By doing so, the sorted list can be directly utilized for rendering (\ie, alpha compositing). Fig.~\ref{fig:fig_3dgs}c and Fig.~\ref{fig:fig_3dgs}d provide the visual demonstration of such concepts. It's worth highlighting that rendering each tile and pixel occurs \textbf{independently}, making this process highly suitable for parallel computations. An additional benefit is that each tile's pixels can access a common \textbf{shared memory} and maintain an \textbf{uniform read sequence} (Fig.~\ref{fig:fig_tile_rendering}), enabling parallel execution of alpha compositing with increased efficiency. In the official implementation of the original paper~\cite{kerbl20233d}, the framework regards the processing of tiles and pixels as analogous to the blocks and threads, respectively, in CUDA programming architecture.

\subsection{Optimization of 3D Gaussian Splatting}
\label{sec:3dgs_optm}
At the heart of 3D GS lies an optimization procedure devised to construct a copious collection of 3D Gaussians that accurately captures the scene's essence, thereby facilitating free-viewpoint rendering. On the one hand, the properties of 3D Gaussians should be optimized via differentiable rasterization to fit the textures of a given scene. On the other hand, the number of 3D Gaussians that can represent a given scene well is unknown in advance. We will introduce how to optimize the properties of each Gaussian in Sec.~\ref{sec:3dgs_optm_param} and how to adaptively control the density of the Gaussians in Sec.~\ref{sec:3dgs_optm_dens}. The two procedures are interleaved within the optimization workflow. Since there are many manually set hyperparameters in the optimization process, we omit the notations of most hyperparameters for clarity.

\subsubsection{Parameter Optimization}
\label{sec:3dgs_optm_param}
  
$\bullet$~\textbf{Loss Function.} Once the synthesis of the image is completed, the difference between the rendered image and ground truth can be measured. All the learnable parameters are optimized by stochastic gradient descent using the $\ell_1$ and D-SSIM loss functions:
\begin{equation}
    \mathcal{L} = (1-\lambda)\mathcal{L}_1 + \lambda\mathcal{L}_{\text{D-SSIM}},
\end{equation}
where $\lambda \in [0,1]$ is a weighting factor. Here, $\mathcal{L}_{\text{D-SSIM}}$ is formulated as $1 - \text{SSIM}$, derived from the structural similarity index measure (SSIM). While $\mathcal{L}_1$ minimizes absolute pixel-wise differences, $\mathcal{L}_{\text{D-SSIM}}$ acts as a perceptual metric that enforces structural consistency, ensuring that the synthesized images preserve high-frequency details and local luminance patterns often missed by $\ell_1$ alone.

\noindent$\bullet$~\textbf{Parameter Update.}
Most properties of a 3D Gaussian can be optimized directly through back-propagation. It is essential to note that directly optimizing the covariance matrix $\bm{\Sigma}$ can result in a non-positive semi-definite matrix, which would not adhere to the physical interpretation typically associated with covariance matrices. To circumvent this issue, 3D GS chooses to optimize a quaternion $\bm{q}$ and a 3D vector $\bm{s}$. Here $\bm{q}$ and $\bm{s}$ represent rotation and scale, respectively. This approach allows the covariance matrix $\bm{\Sigma}$ to be reconstructed as follows:
\begin{equation}
    \bm{\Sigma} = \bm{RS}\bm{S}^{\top}\bm{R}^{\top},
\end{equation}
where $\bm{R}$ is the rotation matrix derived from the quaternion $\bm{q}$, and $\bm{S}$ is the scaling matrix given by $\text{diag}(\bm{s})$. As seen, there is a complex computational graph to obtain the opacity $\alpha$, \ie, $\bm{q}$ and $\bm{s} \mapsto \bm{\Sigma}$, $ \bm{\Sigma} \mapsto \bm{\Sigma}'$, and $ \bm{\Sigma}' \mapsto \alpha$. To avoid the cost of automatic differentiation, 3D GS derives the gradients for $\bm{q}$ and $\bm{s}$ to compute them directly during optimization.

\subsubsection{Density Control}
\label{sec:3dgs_optm_dens}

\noindent$\bullet$~\textbf{Initialization.} 3D GS starts with the initial set of sparse points from SfM or random initialization. Note that a good initialization is essential to convergence and reconstruction quality~\cite{cheng2024gaussianpro}. Afterwards, point densification and pruning are adopted to control the density of 3D Gaussians.

\noindent$\bullet$~\textbf{Point Densification.}
In the point densification phase, 3D GS adaptively increases the density of Gaussians to better capture the details of a scene. This process focuses on areas with missing geometric features or regions where Gaussians are too spread out. The densification procedure will be performed at regular intervals (\ie, after a certain number of training iterations), focusing on those Gaussians with large view-space positional gradients (\ie, above a specific threshold). It involves either cloning small Gaussians in under-reconstructed areas or splitting large Gaussians in over-reconstructed regions. For cloning, a copy of the Gaussian is created and moved towards the positional gradient. For splitting, a large Gaussian is replaced with two smaller ones, reducing their scale by a specific factor. This step seeks an optimal distribution and representation of Gaussians in 3D space, enhancing the overall quality of the reconstruction.

\noindent$\bullet$~\textbf{Point Pruning.}
The point pruning stage involves the removal of superfluous or less impactful Gaussians, which can be viewed as a regularization process. It is executed by eliminating Gaussians that are virtually transparent (with $\alpha$ below a specified threshold) and those that are excessively large in either world-space or view-space. In addition, to prevent unjustified increases in Gaussian density near input cameras, the alpha value of the Gaussians is set close to zero after a certain number of iterations. This allows for a controlled increase in the density of necessary Gaussians while enabling the culling of redundant ones. The process not only helps in conserving computational resources but also ensures that the Gaussians in the model remain precise and effective for the representation of the scene.

\section{3D Gaussian Splatting: Directions}
\label{sec:e3dgs}

Though 3D GS has achieved impressive milestones, significant room for improvement remains, \eg, data and hardware requirement, rendering and optimization algorithm, and applications in downstream tasks. In the subsequent sections, we seek to elaborate on select extended versions. These are: \textbf{i}) 3D GS for Sparse Input (Sec. \ref{sec:e3dgs_de3dgs}), \textbf{ii}) Memory-efficient 3D GS (Sec. \ref{sec:e3dgs_me3dgs}), \textbf{iii}) Photorealistic 3D GS (Sec. \ref{sec:e3dgs_photo}), \textbf{iv}) Improved Optimization Algorithms (Sec. \ref{sec:e3dgs_optim}), \textbf{v}) 3D Gaussian with More Properties (Sec. \ref{sec:e3dgs_property}), \textbf{vi}) Hybrid Representation (Sec. \ref{sec:e3dgs_struct}), and \textbf{vii}) New Rendering Algorithm (Sec. \ref{sec:e3dgs_new_rendering}). To provide a synthesized overview before delving into each direction, Table~\ref{tbl:direction_summary} summarizes their key ideas, representative works (ordered chronologically), advantages, and limitations. While we have carefully selected several key directions, we acknowledge that it is inevitably a biased view. A more comprehensive collection is given in \href{https://github.com/guikunchen/Awesome3DGS}{Github}.
 
\subsection{3D GS for Sparse Input}
\label{sec:e3dgs_de3dgs}

A notable issue of 3D GS is the emergence of artifacts in areas with insufficient observational data. This challenge is a prevalent limitation in radiance field rendering, where sparse data often leads to inaccuracies in reconstruction. From a practical perspective, reconstructing scenes from limited viewpoints is of significant interest, particularly for the potential to enhance functionality with minimal input.

Existing methods can be categorized into two groups (\textit{cf.} Table~\ref{tbl:sparse_tradeoff}). \textbf{i) Regularization} based methods introduce additional constraints such as depth information to enhance the detail and global consistency~\cite{li2024dngaussian,zhu2023fsgs,chen2024mvsplat,zhang2024cor}. For example, DNGaussian~\cite{li2024dngaussian} introduced a depth-regularized scheme to address the challenge of geometry degradation. FSGS~\cite{zhu2023fsgs} devised a Gaussian Unpooling process for initialization and also introduced depth regularization. MVSplat~\cite{chen2024mvsplat} proposed a cost volume representation so as to provide geometry cues. Unfortunately, when dealing with a limited number of views, or even just one, the efficacy of regularization techniques tends to diminish, which leads to \textbf{ii) generalizability} based methods that use learned priors~\cite{charatan2023pixelsplat,szymanowicz2023splatter,xu2024grm,szymanowicz2024flash3d}. One approach involves synthesizing additional views through generative models, which can be seamlessly integrated into existing reconstruction pipelines~\cite{sargent2024zeronvs}. However, this augmentation strategy is computationally intensive and inherently bounded by the capabilities of the used generative model. Another well-known paradigm employs feed-forward Gaussian model to directly generates the properties of a set of 3D Gaussians. This paradigm typically requires multiple views for training but can reconstruct 3D scenes with only one input image. For instance, PixelSplat~\cite{charatan2023pixelsplat} proposed to sample Gaussians from dense probability distributions. Splatter Image~\cite{szymanowicz2023splatter} introduced a 2D image-to-image network that maps an input image to a 3D Gaussian per pixel. However, as the generated pixel-aligned Gaussians are distributed nearly evenly in the space, they struggle to represent high-frequency details and smoother regions with an appropriate number of Gaussians.

The challenge of 3DGS for sparse inputs centers on how priors are modeled, whether through depth information, generative models, or feed-forward Gaussian models. In practice, there is a trade-off between regularization-based methods, which rely on strong constraints and per-scene optimization (favoring geometric quality but incurring higher computation), and generalizability-based methods, which use learned priors for fast feed-forward inference (favoring speed and cross-scene generalization, but sometimes at the expense of geometric fidelity). We summarize these characteristics in Table~\ref{tbl:sparse_tradeoff}. Future research could explore adaptive mechanisms to control this trade-off, for example via learned confidence measures, context-aware prior selection, or user preferences, \etc In addition, while most current methods focus on static scenes, extending these approaches to dynamic scenarios presents an exciting direction, particularly in handling temporal consistency and motion-induced artifacts.

\subsection{Memory-efficient 3D GS}
\label{sec:e3dgs_me3dgs}

While 3D GS demonstrates remarkable capabilities, its scalability poses significant challenges, particularly when juxtaposed with NeRF-based methods. The latter benefits from the simplicity of storing merely the parameters of a learned MLP. This scalability issue becomes acute in the context of large-scale scene management, where the computational and memory demands escalate substantially. Consequently, there is an urgent need to optimize memory usage in both model training and storage. Please see the excellent surveys~\cite{bagdasarian20253dgs,ali2025compression} for more insights on GS compression.

Recent research has pursued two primary directions to address memory efficiency. First, several approaches focus on \textbf{reducing the number of 3D Gaussians}~\cite{papantonakis2024reducing,lee2023compact,chen2024hac}. These methods either employ strategic pruning of low-impact Gaussians, such as the volume-based masking~\cite{lee2023compact}, or represent neighboring Gaussians using the same properties stored within a ``local anchor'' obtained by clustering~\cite{lu2023scaffold}, hash-grid~\cite{chen2024hac}, \etc Second, researchers have developed methods for \textbf{compressing Gaussian's properties}~\cite{niedermayr2024compressed,lee2023compact,chen2024hac}. For instance, Niedermayr \etal~\cite{niedermayr2024compressed} compressed color and Gaussian parameters into compact codebooks, using sensitivity measures for effective quantization and fine-tuning. HAC~\cite{chen2024hac} predicted the probability of each quantized attribute using Gaussian distributions and then devise an adaptive quantization module. These directions are not mutually exclusive; instead, one framework might use a hybrid approach combining multiple strategies.

While current compression techniques have achieved significant storage reduction ratios (often by factors of 10-20$\times$), several challenges remain. One practical challenge is deploying 3D GS on edge devices, where tight memory budgets can limit how much of a large scene can be stored or loaded locally. The field particularly needs advances in memory efficiency during the training phase, potentially through quantization-aware training protocols, the development of scene-agnostic, reusable codebooks, \etc Furthermore, optimizing the trade-off between compression efficiency and visual fidelity remains an open problem.

\subsection{Photorealistic 3D GS}
\label{sec:e3dgs_photo}

The current rendering pipeline of 3D GS (Sec. \ref{sec:3dgs_nvs}) is straightforward and involves several drawbacks. For instance, the simple visibility algorithm may lead to a drastic switch in the depth/blending order of Gaussians. The visual fidelity of rendered images, including aspects such as aliasing, reflections, and artifacts, can be further optimized.

Recent research has focused on addressing three main aspects of visual quality, with aliasing being specific to 3D GS's rendering algorithm, while reflection and blur handling represent broader challenges in 3D reconstruction. \textbf{i) Aliasing}. Due to the discrete sampling paradigm (viewing each pixel as a single point instead of an area), 3D GS is susceptible to aliasing when dealing with varying resolutions, which leads to blurring or jagged edges. Solutions emerged at both training and inference stages. Researchers developed training-time improvements from the sampling rate perspective and introduced schemes such as multi-scale Gaussians~\cite{yan2023multi}, 2D Mip filter~\cite{yu2023mip} (predecessor of Mip-NeRF~\cite{barron2021mip}), and conditioned logistic function~\cite{liang2024analytic}. Inference-time solutions, such as 2D scale-adaptive filtering~\cite{song2024sa}, offer enhanced fidelity that can be integrated into any existing 3D GS frameworks. \textbf{ii) Challenging objects}. Realistic reconstruction of objects with complex appearance, such as reflective, transparent, highly glossy, or anisotropic surfaces, remains a long-standing problem. 3D GS inherits these difficulties because its rasterization-based pipeline only approximates secondary effects. Recent works therefore extend 3D GS with shading functions defined on per-Gaussian normals to better handle reflective and specular surfaces~\cite{jiang2023gaussianshader}, with mirror attributes and plane-mirror imaging to reconstruct mirror geometry and reflections~\cite{meng2024mirror}, and with anisotropic view-dependent appearance fields that more faithfully capture specular and anisotropic components~\cite{yang2024spec}, as well as relightable Gaussian representation~\cite{saito2023relightable}. Despite these advances, robust reconstruction of reflective and transparent objects in general scenes remains challenging. \textbf{iii) Blur}. While 3D GS excels on carefully curated datasets, real-world captures often suffer from blurs such as motion blur and defocus blur. Recent approaches explicitly incorporated blur modeling during training, employing techniques such as coarse-to-fine kernel optimization~\cite{peng2024bags} and photometric bundle adjustment~\cite{zhao2024bad} to address this challenge.

While the approximations made in GS (Sec. \ref{sec:3dgs_nvs}) contribute to its efficiency, they also underlie artifacts such as aliasing and blur, and limit the accuracy of illumination estimation and the modeling of challenging objects. Most existing efforts focus on specific failure modes (\eg, anti-aliasing, deblurring, or handling particular challenging objects) rather than providing a universal solution. In practice, one can first diagnose which issues are present in a scene (\eg, aliasing, blur, or reflective and transparent objects) and then apply targeted optimization strategies. For static object reconstruction in particular, faithfully modeling challenging objects such as reflective, transparent, or highly view-dependent surfaces remains difficult despite recent progress, and extending these capabilities to dynamic or relightable scenes further increases the complexity. The ultimate goal is to develop more general reconstruction systems that alleviate these limitations, either through fundamental improvements to 3D GS or through brand-new architectures.

\subsection{Improved Optimization Algorithms}
\label{sec:e3dgs_optim}

The optimization of 3D GS presents several challenges that affect the quality of reconstruction. These include issues with convergence speed, visual artifacts from improper Gaussians, and the need for better regularization during optimization. The raw optimization method (Sec. \ref{sec:3dgs_optm}) might lead to overreconstruction in some regions while underrepresenting others, resulting in blur and visual inconsistencies.

Three main directions stand out for improving the optimization of 3D GS. \textbf{i) Additional Regularization} (\eg, frequency~\cite{zhang2024fregs} and geometry~\cite{lu2023scaffold,li2024geogaussian}). Geometry-aware approaches have been particularly successful, preserving scene structure through the incorporation of local anchor points~\cite{lu2023scaffold}, depth and surface constraints~\cite{chen2023neusg,yu2024gaussian,zhang2024rade}, Gaussian volumes~\cite{chen2024lara}, \etc \textbf{ii) Optimization Procedure Enhancement}~\cite{cheng2024gaussianpro,yu2024gaussian,ververas2024sags}. While the original strategy of density control (Sec. \ref{sec:3dgs_optm_dens}) has proven valuable, considerable room for improvement remains. For example, GaussianPro~\cite{cheng2024gaussianpro} addresses the challenge of dense initialization in texture-less surfaces and large-scale scenes through an advanced Gaussian densification strategy. \textbf{iii) Constraint Relaxation}. Reliance on external tools/algorithms can introduce errors and cap the system's performance potential. For instance, SfM, commonly used in the initialization process, is error-prone and struggle with complex scenes. Recent works have begun exploring COLMAP-free~\cite{schonberger2016structure} approaches utilizing stream continuity~\cite{fu2023colmap,smith2024flowmap}, potentially enabling learning from internet-scale unposed video datasets. This line of work removes the dependency on pre-computed SfM by jointly optimizing camera poses and Gaussian parameters in an offline/batch manner. Readers interested in the online counterpart of this unknown-pose problem (tracking + mapping under real-time constraints) are referred to Sec.~\ref{sec:app_robotics} on GS-based SLAM.

Though impressive, existing methods primarily concentrate on optimizing Gaussians to accurately reconstruct scenes from scratch, neglecting a challenging yet promising solution which reconstructs scenes in a few-shot manner through established ``meta representations''. Such solution could enable adaptive meta-learning strategies that combine scene-specific and general knowledge. See ``learning physical priors from large-scale data'' in Sec.~\ref{sec:direction} for further insights.

\subsection{3D Gaussian with More Properties}
\label{sec:e3dgs_property}

Despite impressive, the properties of 3D Gaussian (Sec. \ref{sec:3dgs_nvs}) are designed to be used for novel-view synthesis only. By augmenting 3D Gaussian with additional properties, such as linguistic~\cite{shi2023language,qin2023langsplat,zuo2024fmgs}, semantic/instance~\cite{zhou2023feature,ye2023gaussian,cen2023segment}, and spatial-temporal~\cite{yang2023real} properties, 3D GS demonstrates its considerable potential to revolutionize various domains.

Here we list several interesting applications using 3D Gaussians with specially designed properties. \textbf{i}) \textbf{Language Embedded Scene Representation}~\cite{shi2023language,qin2023langsplat,zuo2024fmgs}. Due to the high computational and memory demands of current language-embedded scene representations, Shi \etal~\cite{shi2023language} proposed a quantization scheme that augments 3D Gaussian with streamlined language embeddings instead of the original high-dimensional embeddings. This method also mitigated semantic ambiguity and enhanced the precision of open-vocabulary querying by smoothing out semantic features across different views, guided by uncertainty values. \textbf{ii}) \textbf{Scene Understanding and Editing}~\cite{zhou2023feature,ye2023gaussian,cen2023segment}. Feature 3DGS~\cite{zhou2023feature} integrated 3D GS with feature field distillation from 2D foundation models. By learning a lower-dimensional feature field and applying a lightweight convolutional decoder for upsampling, Feature 3DGS achieved faster training and rendering speeds while enabling high-quality feature field distillation, supporting applications like semantic segmentation and language-guided editing. \textbf{iii}) \textbf{Spatiotemporal Modeling}~\cite{yang2023real,lin2023gaussian} (task-centric summary and comparisons are given in Sec.~\ref{sec:app_dynamic_scene}). To capture the complex spatial and temporal dynamics of 3D scenes, Yang \etal~\cite{yang2023real} conceptualized spacetime as a unified entity and approximates the spatiotemporal volume of dynamic scenes using a collection of 4D Gaussians. The proposed 4D Gaussian representation and corresponding rendering pipeline are capable of modeling arbitrary rotations in space and time and allow for end-to-end training.

\subsection{Hybrid Representation}
\label{sec:e3dgs_struct}
Rather than augmenting 3D Gaussian with additional properties, another promising avenue of adapting to downstream tasks is to introduce structured information (\eg, spatial MLPs and grids) tailored for specific applications.

Next we showcase various fascinating uses of 3D GS with specially devised structured information. \textbf{i}) \textbf{Facial Expression Modeling}. Considering the challenge of creating high-fidelity 3D head avatars under sparse view conditions, Gaussian Head Avatar~\cite{xu2023gaussian} introduced controllable 3D Gaussians and an MLP-based deformation field. Concretely, it captured detailed facial expressions and dynamics by optimizing neutral 3D Gaussians alongside the deformation field, thus ensuring both detail fidelity and expression accuracy. \textbf{ii}) \textbf{Spatiotemporal Modeling} (task-centric summary and comparisons are given in Sec.~\ref{sec:app_dynamic_scene}). Yang \etal~\cite{yang2023deformable} proposed to reconstruct dynamic scenes with deformable 3D Gaussians. The deformable 3D Gaussians are learned in a canonical space, coupled with a deformation field (\ie, a spatial MLP) that models the spatial-temporal dynamics. The proposed method also incorporated an annealing smoothing training mechanism to enhance temporal smoothness without additional computational costs. \textbf{iii}) \textbf{Style Transfer}. Saroha \etal~\cite{saroha2024gaussian} proposed GS in style, an advanced approach for real-time neural scene stylization. To maintain a cohesive stylized appearance across multiple views without compromising on rendering speed, they used pre-trained 3D Gaussians coupled with a multi-resolution hash grid and a small MLP to produce stylized views. In a nutshell, incorporating structured information can serve as a complementary part for adapting to tasks that are incompatible with the sparsity and disorder of 3D Gaussians.

\subsection{New Rendering Algorithm for 3D Gaussians}
\label{sec:e3dgs_new_rendering}

While the rasterization-based pipeline of 3D GS offers impressive real-time performance, it still suffers from the inherent limitations, including inefficient handling of highly-distorted cameras (for robotics), secondary rays (for optical effects like reflections and shadows), and stochastic ray sampling (needed in various existing pipelines). In addition, the assumptions that Gaussians do not overlap and can be sorted accurately using only centers are often violated in practice, leading to temporal artifacts when camera movement changes sorting order. This issue is closely related to the long-standing problem of transparency and visibility ordering in classical rasterization, where order-independent transparency (OIT) methods aim to reduce sorting artifacts. Weighted-blended OIT~\cite{mcguire2013weighted} provides a widely-used solution that approximates transparency without explicit per-pixel sorting.

Recent works~\cite{moenne20243d,mai2024ever,condor2025don} explored ray tracing based rendering algorithms as an alternative. For instance, GaussianTracer~\cite{moenne20243d} introduced a new ray tracing implementation for Gaussian primitives, and devised several accelerating strategies according to the uneven density and interleaved nature of Gaussians. EVER~\cite{mai2024ever} deivsed a physically accurate, constant density ellipsoid representation that allows for the exact computation of the volume rendering integral, rather than relying on somewhat satisfactory approximations. This advancement eliminates popping artifacts.

Thanks to the fundamental paradigm shift, several exciting possibilities might emerge, including advanced optical effects (reflection, refraction, shadows, global illumination, \etc), support for complex camera models (highly-distorted lenses, rolling shutter effects, \etc), physically accurate rendering with true directional appearance evaluation (\vs tile based approximation), and more. While these capabilities currently come with additional computational costs, they provide essential building blocks for future research in inverse rendering, physical material modeling, relighting, and complex scene reconstruction.

\subsection{Kernel Engineering}
\label{sec:kernel_engineering}

Gaussian kernel is intrinsically smooth with infinite spatial support. Consequently, representing sharp boundaries, thin surfaces, and piecewise-planar structures can be parameter-inefficient: capturing high-frequency details requires aggressive densification and careful regularization. Furthermore, overlapping semi-transparent splats can produce cloudy artifacts near discontinuities due to the kernel's low-pass nature and cross-boundary leakage.

To address these limitations, a growing body of work revisits the geometric primitive and the splatting operator. These efforts can be broadly categorized as follows:
\textbf{i) Surface-aligned primitives}~\cite{huang20242d,dai2024high} constrain splatted elements to align with local tangent planes. By utilizing planar or near-planar parameterizations alongside geometric regularizers, this direction explicitly forces primitives to match the underlying surface geometry, thereby improving multi-view consistency and the quality of extracted meshes.
\textbf{ii) Alternative distributions and shape modifications}~\cite{li20253d,huang2025deformable,zhu20253d,liu2025deformable,liu2025universal} alter the mathematical formulation and spatial footprint of the kernel to handle sharp transitions. By replacing the standard Gaussian with bounded-support Beta distributions, asymmetric cuts, deformable radial bases, or heavy-tailed mixtures with subtractive components (\eg, scooping), these methods provide finer control over local attenuation. This prevents cross-boundary density leakage and allows for a more compact representation of complex frequency profiles and hard edges.
\textbf{iii) Non-Gaussian geometric primitives and operator redesigns}~\cite{held20253d,franke2024trips,ye2025gaussian,condor2025don} abandon the standard ellipsoid in favor of alternative shapes, or fundamentally modify the rendering operator (Sec.~\ref{sec:e3dgs_new_rendering}). This includes utilizing 3D smooth convexes, trilinear point splatting into image pyramids, or bi-scale pipelines that combine standard graphics rasterization with sorting-free splatting. These redesigns natively accommodate distinct geometric features and bypass the limitations of pure Gaussian rasterization.

Despite rapid progress, kernel engineering introduces new practical trade-offs. Alternative parameterizations consistently require custom rendering implementations and tailored primitive management strategies (\eg, modified splitting, merging, and pruning heuristics). Furthermore, standardized evaluation remains challenging, as comparisons are frequently confounded by differences in primitive budgets. As early hybrid pipelines begin to emerge, a promising direction lies in the principled compression and adaptive allocation of heterogeneous primitive types -- deploying computationally complex kernels only where specific geometric intricacies demand them.

\section{Application Areas and Tasks}
\label{sec:application}
\begin{figure*}[!t]
\centering
\includegraphics[width=1.0\textwidth]{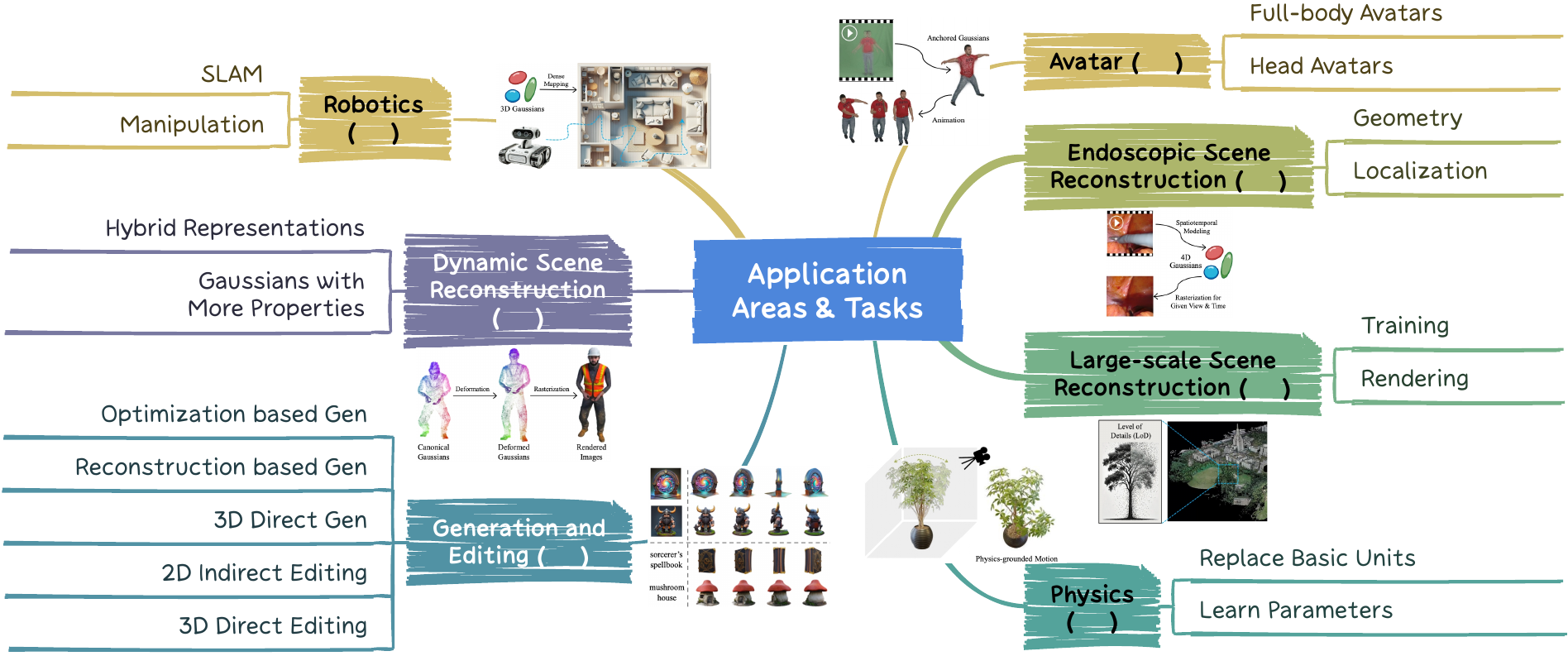}
\put(-332, 140.5){\small{\ref{sec:app_robotics}}}
\put(-292.5, 90.3){\small{\ref{sec:app_dynamic_scene}}}
\put(-280.2, 25){\small{\ref{sec:app_gen_edit}}}
\put(-117.2, 160){\small{\ref{sec:app_avatar}}}
\put(-89, 127.3){\small{\ref{sec:app_endo}}}
\put(-87.8, 70.3){\small{\ref{sec:app_largescale}}}
\put(-135.1, 6.8){\small{\ref{sec:app_physics}}}
\put(-427, 158){\footnotesize{\cite{yan2023gs,keetha2023splatam,matsuki2023gaussian}}}
\put(-427, 143.7){\footnotesize{\cite{lu2024manigaussian,shorinwa2024splat}}}
\put(-427, 115){\footnotesize{\cite{yang2023deformable}}}
\put(-427, 97){\footnotesize{\cite{luiten2023dynamic,yang2023real}}}
\put(-429, 64){\footnotesize{\cite{tang2023dreamgaussian}}}
\put(-429, 50){\footnotesize{\cite{tang2024lgm}}}
\put(-429, 35){\footnotesize{\cite{mu2024gsd,he2024gvgen,pan2024humansplat}}}
\put(-429, 20.5){\footnotesize{\cite{chen2023gaussianeditor,fang2023gaussianeditor,palandra2024gsedit}}}
\put(-429, 6){\footnotesize{\cite{huang2023sc,waczynska2024games,gao2024real}}}
\put(-30, 174){\footnotesize{\cite{li2023animatable,lei2023gart}}}
\put(-42.5, 159.3){\footnotesize{\cite{qian2023gaussianavatars,zhou2024headstudio,li2024talkinggaussian}}}
\put(-26, 145.3){\footnotesize{\cite{huang2024endo,zhao2024hfgs}}}
\put(-16.2, 130.3){\footnotesize{\cite{wang2024endogslam}}}
\put(-29, 87.8){\footnotesize{\cite{liu2024citygaussian,lin2024vastgaussian}}}
\put(-16, 73.6){\footnotesize{\cite{kerbl2024hierarchical}}}
\put(-36, 24.7){\footnotesize{\cite{xie2023physgaussian,feng2024gaussian,borycki2024gasp}}}
\put(-39, 10.2){\footnotesize{\cite{huang2024dreamphysics,zhang2024physdreamer,zhong2024reconstruction}}}

\vspace{-1em} 
\caption{Typical applications benefited from GS (Sec.~\ref{sec:application}). Some images are borrowed from~\cite{lu20243d,tang2024lgm,zhu2024deformable,xiong2024gauu,lei2023gart,xie2023physgaussian} and redrawn.}
\label{fig:app}
\vspace{-1em} 
\end{figure*}

Building on the rapid advancements in 3D GS, a wide range of innovative applications has emerged across multiple domains (\cf Fig.~\ref{fig:app}) such as robotics (Sec.~\ref{sec:app_robotics}), dynamic scene reconstruction and representation (Sec.~\ref{sec:app_dynamic_scene}), generation and editing (Sec.~\ref{sec:app_gen_edit}), avatar (Sec.~\ref{sec:app_avatar}), medical systems (Sec.~\ref{sec:app_endo}), large-scale scene reconstruction (Sec.~\ref{sec:app_largescale}), physics (Sec.~\ref{sec:app_physics}), and even other scientific disciplines~\cite{zhang2024togs,wu2024dual,zhang2024darkgs,li2024ggrt}. Here, we highlight key examples that underscore the transformative impact and potential of 3D GS and offer a more comprehensive collection in \href{https://github.com/guikunchen/Awesome3DGS}{Github}. We also provide a collection of representative datasets (\cf Table~\ref{tbl:dataset}) according to our taxonomy.

\subsection{Robotics}
\label{sec:app_robotics}

The evolution of scene representation in robotics has been profoundly shaped by NeRF. However, NeRF's computational cost poses a critical bottleneck for real-time applications. The shift from implicit to explicit representation not only accelerates optimization but also unlocks direct access to spatial and structural scene data, making 3D GS a transformative tool for robotics. Its ability to balance high-fidelity reconstruction with computational efficiency positions 3D GS as a cornerstone for advancing robotic perception, manipulation, and navigation in real-world environments.

The integration of GS into robotic systems has yielded advancements across three core domains. In \textbf{SLAM}, GS-based methods~\cite{huang2023photo,matsuki2023gaussian,yugay2023gaussian,yan2023gs,keetha2023splatam,hong2024liv,ji2024neds,sun2024high,tosi2024nerfs,li2024sgs,zhu2024semgauss,deng2024compact,gao20253d,gao2026uncertainty,feng2025gaussian,hu2024cg,lang2024gaussian,peng2024rtg} excel in real-time dense mapping but face inherent trade-offs. Visual SLAM frameworks, particularly RGB-D variants~\cite{keetha2023splatam,yugay2023gaussian,sun2024high}, leverage depth supervision for geometric fidelity but falter in low-texture or motion-degraded environments. RGB-only approaches~\cite{matsuki2023gaussian,huang2023photo,sandstrom2024splat} circumvent depth sensors but grapple with scale ambiguity and drift. Multi-sensor fusion strategies, such as LiDAR integration~\cite{hong2024liv,lang2024gaussian,wu2024mm}, enhance robustness in unstructured settings at the cost of calibration complexity. Semantic SLAM~\cite{zhu2024semgauss,ji2024neds,li2024sgs,yanopen} extends scene understanding through object-level semantics but struggles with scalability due to lighting sensitivity in color-based methods or computational overhead in feature-based methods. GS-based SLAM can be viewed as the online counterpart of pose-free reconstruction in Sec.~\ref{sec:e3dgs_optim}: both address unknown poses by jointly optimizing camera trajectories and the Gaussian map, but SLAM operates sequentially with real-time constraints and additional mechanisms to control drift (\eg, keyframe selection and loop closure).
3D GS based \textbf{manipulation}~\cite{lu2024manigaussian,abou2024physically,shorinwa2024splat,ji2024graspsplats,zheng2024gaussiangrasper} bypasses the need for auxiliary pose estimation in NeRF-based methods, enabling rapid single-stage tasks like grasping in static environments via geometric and semantic attributes encoded in Gaussian properties. Multi-stage manipulation~\cite{lu2024manigaussian,shorinwa2024splat}, where environmental dynamics demand real-time map updates, requires explicit modeling of dynamic adjustments (\eg, object motions and interactions), material compliance, \etc

The advancement of 3D GS in robotics faces three pivotal challenges. First, adaptability in dynamic and unstructured environments remains critical: real-world scenes are rarely static, requiring systems to continuously update representations amid motion, occlusions, and sensor noise without sacrificing accuracy. Second, current semantic mapping methods rely on costly, scene-specific optimization processes, limiting generalizability and scalability for real-world deployment.  Third, unlike NeRF based systems which can use MLP parameters as input features for downstream decision-making, 3D Gaussians' inherent lack of spatial order complicates feature aggregation, with no standardized framework yet established. Bridging the gap between high-fidelity reconstruction and actionable semantic/physical understanding will define the next frontier for 3D GS, moving beyond passive mapping towards embodied intelligence.

\subsection{Dynamic Scene Reconstruction}
\label{sec:app_dynamic_scene}

Dynamic scene reconstruction refers to the process of capturing and representing the three-dimensional structure and appearance of a scene that changes over time~\cite{pumarola2021d,park2021hypernerf,park2021nerfies,guo2023forward}. This involves creating a digital model that accurately reflects the geometry, motion, and visual aspects of the objects in the scene as they evolve. Dynamic scene reconstruction is crucial in various applications, \eg, VR/AR, 3D animation, and autonomous driving~\cite{zhou2023drivinggaussian,yan2024street,zhou2024hugs}.

The key to adapt 3D GS to dynamic scenes is the modeling of temporal dimension which allows for the representation of scenes that change over time. 3D GS based methods~\cite{yang2023deformable,yang2023real,lin2023gaussian,kratimenos2023dynmf,das2023neural,shaw2023swags,wu20234d,luiten2023dynamic,huang2023sc,shao2023control4d,yu2023cogs,liang2023gaufre,katsumata2023efficient,li2023spacetime,guo2024motion,bae2024per,lei2024mosca,wang2024shape,duan20244d} for dynamic scene reconstruction can generally be divided into two main categories as discussed in Sec.~\ref{sec:e3dgs_property} and Sec.~\ref{sec:e3dgs_struct}. The first category utilizes \textbf{additional fields} like spatial MLPs or grids to \textbf{model deformation} (Sec.~\ref{sec:e3dgs_struct}). For example, Yang \etal~\cite{yang2023deformable} first proposed deformable 3D Gaussians tailored for dynamic scenes. These 3D Gaussians are learned in a canonical space and can be used to model spatial-temporal deformation with an implicit deformation field (implemented as an MLP). GaGS~\cite{lu20243d} devised the voxelization of a set of Gaussian distributions, followed by the use of sparse convolutions to extract geometry-aware features, which are then utilized for deformation learning.  On the other hand, the second category is based on the idea that scene changes can be \textbf{encoded into the 3D Gaussian representation} with a specially designed rendering process (Sec.~\ref{sec:e3dgs_property}). For instance, Luiten \etal~\cite{luiten2023dynamic} introduced dynamic 3D Gaussians to model dynamic scenes by keeping the properties of 3D Gaussians unchanged over time while allowing their positions and orientations to change. Yang \etal~\cite{yang2023real} designed a 4D Gaussian representation, where additional properties are used to represent 4D rotations and spherindrical harmonics, to approximate the spatial-temporal volume of scenes.

While 3D GS advances dynamic scene reconstruction by modeling per-Gaussian deformations, its reliance on fine-grained primitives limits scalability and robustness. Current methods struggle to balance computational efficiency and precision: small-scale reconstructions unify dynamic and static elements but become intractable in large environments, often requiring manual priors to segment regions -- a barrier in unstructured settings. Furthermore, the absence of object-level motion reasoning leads to artifacts and poor generalization over long sequences. Future work might prioritize object-centric frameworks that group Gaussians into persistent entities so as to model inherent motion disentanglement (dynamic \vs static).

\subsection{Generation and Editing}
\label{sec:app_gen_edit}

Content generation and editing are two closely related capabilities in 3D creation pipelines. Generation aims to synthesize new assets or scenes from high-level conditions such as text and images, while editing focuses on modifying existing content in a controlled way. In this subsection, we concentrate on works that take 3D GS as the target representation. Please see the excellent survey~\cite{he2025survey} for more insights on the GS based generation and editing methods.

$\bullet$~\textbf{Generation.} Recent advances in GS based generation~\cite{chen2023text,tang2023dreamgaussian,yi2023gaussiandreamer,liang2023luciddreamer,liu2023humangaussian,yang2023learn,zou2023triplane,ling2023align,ren2023dreamgaussian4d,yin20234dgen,zhang2023repaint123,pan2024fast,xu2024agg,tang2024lgm,yang2024gaussianobject,barthel2024gaussian,jiang2024brightdreamer,zhou2024dreamscene360,zhuo2024vivid,wu2024sc4d,he2024gvgen,yang2024hash3d,kim2024synctweedies,feng2024fdgaussian,li2024dreamscene,melas20243d,li2024controllable,zhang2024gaussiancube,mu2024gsd,lee2024vividdream} can be roughly grouped along the underlying learning paradigm. \textbf{Optimization}-based methods follow the score-distillation paradigm: a powerful 2D or multi-view diffusion model is queried from many camera poses, and its gradients are back-propagated through the differentiable GS renderer to update Gaussian parameters. Early text- and image-to-3D pipelines such as DreamGaussian~\cite{tang2023dreamgaussian} and GaussianDreamer~\cite{yi2023gaussiandreamer} treat each scene independently and directly optimize positions, scales, colors, and opacities to match diffusion-model guidance while respecting RGB reconstruction losses. This per-scene fitting often yields high-fidelity geometry and appearance for object- and scene-level content, and can be naturally extended to dynamic or 4D generation with temporal regularization. However, it incurs heavy computational cost (minutes to hours per prompt) and is prone to Janus ambiguity, over-smoothing, or texture oversaturation unless additional regularizers, multi-view diffusion priors, or geometry-aware constraints are introduced. \textbf{Reconstruction}-based methods decompose the problem into two classical stages: multi-view generation followed by 3D reconstruction. Given a text or image prompt, a pre-trained multi-view or video diffusion model first synthesizes a small set of posed views (and sometimes depth or camera trajectories). A GS reconstructor such as LGM~\cite{tang2024lgm} or VividDream~\cite{lee2024vividdream} then treats these images as supervision and solves a standard multi-view reconstruction problem to obtain Gaussians. This design reuses mature reconstruction pipelines and makes system-level engineering straightforward, but inherits the fundamental limitations of the multi-view generator: imperfect cross-view consistency can lead to noisy surface geometry, duplicated structures, and degraded textures, especially in occluded or textureless regions. \textbf{Direct 3D generation} methods aim to learn a feed-forward or diffusion model that outputs Gaussian fields directly from text or images, without an explicit intermediate image set~\cite{mu2024gsd,he2024gvgen,zhang2024gaussiancube,pan2024humansplat,cai2025baking}. Trained on large collections of synthetic or reconstructed data, these models amortize the reconstruction cost and naturally promote multi-view coherence, enabling near-real-time sampling at test time. The main challenges lie in the high-dimensional and highly structured output space (positions, anisotropic covariance, colors, opacity), the need for large, curated 3D training corpora, and maintaining controllability (\eg, disentangling geometry from appearance) while scaling to diverse objects and scenes. Scene-level versions further have to reason about layout, occlusions, and long-range dependencies. Beyond applications, these methods reflect a paradigm shift from per-scene optimization of Gaussians to amortized, feed-forward prediction (often with generative priors), enabling much lower test-time latency and better scalability (Sec.~\ref{sec:direction}).

$\bullet$~\textbf{Editing.} Current editing works~\cite{chen2023gaussianeditor,fang2023gaussianeditor,huang2023point,ye2023gaussian,cen2023segment,zhou2023feature,lan20232d,huang2023sc,shao2023control4d,yu2023cogs,zhuang2024tip,dou2024cosseggaussians,hu2024semantic,palandra2024gsedit,gu2024egolifter,lyu2024gaga,liu2024infusion,qiu2024feature,zhang2024stylizedgs,wang2024gscream,zhang20243ditscene,wu2024gaussctrl} can be broadly divided into two families. The first leverages \textbf{2D image-editing} models (e.g., text- or image-conditioned diffusion editors) to drive updates of 3D Gaussians. Representative systems such as GaussianEditor~\cite{chen2023gaussianeditor,fang2023gaussianeditor,palandra2024gsedit} repeatedly render GS views, apply 2D edits, and back-propagate the differences to Gaussian parameters, enabling text- or mask-driven appearance and geometry changes. Follow-up works improve spatial localization and multi-view consistency via attention, view correspondence, or iterative refinement~\cite{wang2024gscream,zhang20243ditscene,wu2024gaussctrl}, and specialize to style transfer or personalization~\cite{zhuang2024tip,zhang2024stylizedgs}. While intuitive and compatible with powerful 2D priors, these pipelines remain computationally heavy and can still suffer from unintended deformations due to weak 3D geometry priors. The second family exploits the explicit, point-based nature of 3D GS for \textbf{direct manipulation}. Semantic or feature-aware approaches enrich Gaussians with labels or embeddings so that users can select and edit subsets of points with text queries, masks, or keypoints~\cite{cen2023segment,ye2023gaussian,hu2024semantic,qiu2024feature,huang2023sc}. Mesh-coupled methods provide stronger structural control. GaMeS~\cite{waczynska2024games} introduces a hybrid ``Gaussian Mesh Splatting'' representation where each Gaussian is parameterized by the vertices of a mesh face; given either an input mesh or an estimated pseudo-mesh, classical mesh operations or animations induce coherent changes in Gaussian position, scale, and orientation. GaussianMesh~\cite{gao2024real} similarly defines Gaussians over an explicit mesh and ties the two via joint refinement: GS rendering guides adaptive face and Gaussian splitting, while mesh constraints regularize Gaussian shapes and support large, interactive deformations driven by mesh editing. However, this family faces several practical challenges: it relies on accurate semantic labels or mesh proxies, so errors in segmentation or remeshing can easily produce artifacts, and editing attributes (\eg, texture, geometry, and illumination) still requires careful regularization and cross-view alignment to preserve plausibility.

\subsection{Avatar}
\label{sec:app_avatar}

Avatars, the digital representations of users in virtual spaces, bridge physical and digital realms, enabling immersive interaction, identity expression, \etc Spanning entertainment (gaming, virtual influencers), enterprise (AI agents, virtual meetings), healthcare, and education, they underpin metaverse economies. Advances in AR and VR amplify their role in redefining social, industrial, and creative landscapes.

3D GS has emerged as a powerful tool for avatar reconstruction, primarily advancing along two directions: full-body modeling and head-centric modeling. For \textbf{full-body avatars}~\cite{jena2023splatarmor,ye2023animatable,li2023animatable,lei2023gart,moreau2023human,kocabas2023hugs,abdal2023gaussian,zheng2023gps,hu2023gauhuman,jiang2023hifi4g,hu2023gaussianavatar,pang2023ash,qian20233dgs,yuan2023gavatar,jung2023deformable,li2023human101,li2024gaussianbody,zhai2025taga}, the current methods typically anchor 3D Gaussians in a canonical space and deform them via parametric body models or cage-based rigging to model dynamic motions. These approaches adopt a hybrid deformation strategy: linear blend skinning handles rigid skeletal transformations such as joint rotations, while pose-conditioned deformation fields account for secondary non-rigid effects like muscle jiggles. For \textbf{head avatars}~\cite{qian2023gaussianavatars,dhamo2023headgas,xiang2023flashavatar,saito2023relightable,chen2023monogaussianavatar,zhao2024psavatar,zhou2024headstudio,li2024talkinggaussian,rivero2024rig3dgs}, the emphasis shifts to modeling intricate facial expressions, fine-grained geometry (e.g., wrinkles, hair~\cite{luo2024gaussianhair}), and dynamic speech-driven animations. Techniques mainly combine parametric morphable face models (\eg, FLAME) with deformable 3D Gaussians, employing diffusion strategies and expression-aware deformation fields to disentangle rigid head poses from non-rigid facial movements. Both directions exploit the speed advantage and editability of GS to enable fast training, real-time rendering, and precise control over deformations, while addressing challenges in cross-frame correspondence, topology flexibility, and multi-view consistency.

Reconstruction in challenging scenes (\eg, occlusions, sparse single-view inputs, or loose clothing) and enhancing avatar interactivity represent critical challenges and opportunities. Parametric model-free methods, which bypass predefined priors by learning skinning weights directly from data, show promise for such scenarios. Complementary to this, generative models can mitigate ambiguities inherent in underconstrained settings. Further integrating physics-based constraints might bridge the gap between static reconstructions and responsive, lifelike interactions, unlocking applications in metaverse, embodied AI, \etc

\subsection{Endoscopic Scene Reconstruction}
\label{sec:app_endo}
Endoscopic Scene reconstruction is pivotal in robot-assisted minimally invasive surgery, enhancing intraoperative navigation, planning, and simulation. Recent advances integrate dynamic radiance fields to address challenges like instrument occlusions and sparse viewpoints in endoscopic videos. However, achieving high fidelity in tissue deformation and topological variation, alongside real-time rendering for latency-sensitive applications, remains critical~\cite{huang2024endo,liu2024endogaussian,zhu2024deformable}.

This task introduces distinct challenges compared to general dynamic scenes, including sparse training data from limited camera mobility in narrow cavities, tool occlusions obscuring critical regions, and single-view geometry ambiguities. Existing approaches mainly used additional depth guidance to infer the geometry of tissues~\cite{huang2024endo,liu2024endogaussian,zhu2024deformable}. For instance, EndoGS~\cite{zhu2024deformable} integrated depth-guided supervision with spatial-temporal weight masks and surface-aligned regularization terms to enhance the quality and speed of 3D tissue rendering while addressing tool occlusion. EndoGaussian~\cite{liu2024endogaussian} introduced two new strategies: holistic Gaussian initialization for dense initialization and spatiotemporal Gaussian tracking for modeling surface dynamics. Zhao \etal~\cite{zhao2024hfgs} argued that these methods suffer from under-reconstruction and proposed to alleviate this problem from frequency perspectives. In addition, EndoGSLAM~\cite{wang2024endogslam} and Gaussian Pancake~\cite{bonilla2024gaussian} devised SLAM systems for endoscopic scenes and showed significant speed advantages.

Advancing endoscopic 3D reconstruction requires targeted efforts in both data and dynamics modeling. Data limitations arise from single-viewpoint videos, which produce ill-posed reconstruction problems due to instrument occlusions and constrained camera mobility, leaving critical tissue regions unobserved. While depth estimators provide temporary workarounds, integrating multi-view camera systems addresses the root cause. In addition, existing datasets often feature truncated sequences (\eg, $4\!\sim\!8s$ in EndoNeRF~\cite{wang2022neural}), which fail to capture prolonged tissue deformation dynamics or complex surgical workflows. Extending temporal coverage to include longer, clinically representative sequences would benefit downstream applications as aforementioned. Modeling limitations persist in current methods, which often represent tissue dynamics at the Gaussian level rather than object- or 3D region-level. This reduces their capacity to encode semantically meaningful anatomical interactions and deserves further explorations.

\subsection{Large-scale Scene Reconstruction}
\label{sec:app_largescale}
Large-scale scene reconstruction is a critical component in fields such as autonomous driving and aerial surveying. Before the emergence of 3D GS, the task has been approached using NeRF based methods, which, while effective for smaller scenes, often fall short in detail and rendering speed when scaled to larger areas (\eg, over 1.5 $km^{2}$). Though GS has demonstrated considerable advantages over NeRFs, the direct application of GS to large-scale environments introduces significant challenges. GS requires an immense number of Gaussians to maintain visual quality, leading to prohibitive GPU memory demands and computational burdens. For instance, a scene spanning 2.7 $km^{2}$ may require over 20 million Gaussians, pushing the limits of commonly used GPUs (\eg, NVIDIA A100 40GB)~\cite{liu2024citygaussian}.

To address the highlighted challenges, researchers have made significant strides in two key areas: \textbf{i}) For \textbf{training}, a divide-and-conquer strategy~\cite{liu2024citygaussian,lin2024vastgaussian,ren2024octree,kerbl2024hierarchical} has been adopted, which segments a large scene into multiple, independent cells. This facilitates parallel optimization for expansive environments. An additional challenge lies in maintaining visual quality, as large-scale scenes often feature texture-less surfaces that can hamper the effectiveness of optimization such as Gaussian initialization and density control (Sec.~\ref{sec:3dgs_optm}). Enhancing the optimization algorithm presents a viable solution to mitigate this issue~\cite{cheng2024gaussianpro,lin2024vastgaussian}. \textbf{ii}) Regarding \textbf{rendering}, the adoption of the Level of Details (LoD) technique from computer graphics has proven instrumental. LoD adjusts the complexity of 3D scenes to balance visual quality with computational efficiency. Current implementations involve feeding only the essential Gaussians to the rasterizer~\cite{lin2024vastgaussian}, or designing explicit LoD structures like the Octree~\cite{ren2024octree} and hierarchy~\cite{kerbl2024hierarchical}. Furthermore, integrating extra input modalities like LiDAR can further enhanced the reconstruction process~\cite{wu2024hgs,wu2024mm,xiong2024gauu}.

One prominent challenge lies in handling sparse or incomplete capture data, which can be mitigated through few-shot adaptation schemes (see Sec.~\ref{sec:e3dgs_de3dgs}) or generalizable priors (see ``learning physical priors from large-scale data'' in Sec.~\ref{sec:direction}). Meanwhile, memory and computational bottlenecks can be addressed via distributed learning strategies~\cite{zhao2024scaling}, such as parameter partitioning across GPU clusters and parallel batched multi-view optimization.

\subsection{Physics}
\label{sec:app_physics}

The simulation of complex real-world dynamics, such as seed dispersal, fracture, or fluid motion, is pivotal for applications spanning VR, animation, scientific modeling, and emerging ``world models'', where usefulness hinges not only on visual plausibility but also on adherence to physical laws. While recent diffusion-based 4D generation frameworks can synthesize visually compelling dynamics, they often optimize for appearance metrics and may violate basic conservation principles or material constraints. In this context, 3D GS provides an attractive bridge between graphics and simulation: Gaussians can be interpreted simultaneously as rendering primitives and as particles or elements in a numerical solver, making it natural to embed physical priors directly into the scene representation.

Existing methods differ mainly in how they couple Gaussians with explicit simulators. One line of work adopts classical schemes such as the material point method (MPM)~\cite{hu2018moving} and position-based dynamics (PBD)~\cite{muller2007position}, and replaces traditional grid or mesh discretizations with 3D Gaussians. In PhysGaussian~\cite{xie2023physgaussian}, for instance, the same Gaussian set is used for both continuum-mechanics-based time integration and for splatting, so physical states evolve on exactly what is rendered. PBD-style pipelines such as Gaussian Splashing~\cite{feng2024gaussian} treat Gaussians as particles subject to constraints, enabling two-way coupling between deformable solids and fluids, while other systems like GASP~\cite{borycki2024gasp} derive mesh-like parameterizations from Gaussians to interface with conventional solvers. These approaches achieve high-fidelity dynamics within their target regimes but inherit the computational cost, stability issues, and parameter tuning burden of the underlying simulators. A complementary direction focuses on \emph{learning} material parameters or effective constitutive models from data while keeping Gaussians as the canonical representation. Physics3D~\cite{liu2024physics3d} builds a viscoelastic MPM on top of 3DGS and estimates elastic and plastic properties by distilling priors from video diffusion models. DreamPhysics~\cite{huang2024dreamphysics}, PhysDreamer~\cite{zhang2024physdreamer}, PhysMotion~\cite{jiang2024vr}, and Feature Splatting~\cite{qiu2024feature} similarly use generative video models as weak teachers for dynamics or material behavior, whereas mass-spring formulations such as Spring-Gaus~\cite{zhong2024reconstruction} encode stiffness and damping explicitly via springs attached to Gaussians. Most frameworks treat Gaussians as discrete particles inside differentiable simulators (with one exception of continuous formulations such as GausSim~\cite{shao2024gausim}), and backpropagate through simulation and rendering to jointly adjust appearance and physics.

Despite rapid progress, physics-based GS is still far from a general-purpose physical world model. Current systems are often specialized to narrow regimes (\eg, a limited set of elastic behaviors), and extending a single Gaussian-based formulation to simultaneously handle rigid, elastic, plastic, granular, and fluid dynamics remains open. Interactions are typically restricted to a few objects in controlled scenes; robust multi-contact, multi-material interactions in cluttered environments still rely on manual priors, and long-horizon rollouts accumulate numerical and modeling errors. Moreover, physical supervision is weak: diffusion-based teachers target perceptual realism rather than quantitative accuracy, and existing datasets rarely provide ground-truth material parameters or stress/strain measurements. Bridging these gaps will require tighter integration between adaptive physics engines and Gaussian representations, better benchmarks for physical correctness, and learning schemes that combine explicit simulation with large-scale data-driven priors.

\section{Performance Comparison}
\label{sec:performance}

In this section, we provide empirical evidence by presenting the performance of several 3D GS algorithms previously discussed. The diverse applications of 3D GS across numerous tasks, coupled with the custom-tailored algorithmic designs for each task, render a uniform comparison of all 3D GS algorithms across a single task or dataset impracticable. For comprehensiveness, we provide a collection of representative datasets in Table~\ref{tbl:dataset} Following the taxonomy outlined Fig.~\ref{fig_structure}. Due to the limited space, we have chosen several representative tasks for an in-depth performance evaluation. The performance scores are primarily sourced from the original papers, except where indicated otherwise.

Quantitative results (\eg, PSNR) can vary noticeably across implementations and hyperparameters (\eg, densification schedules and pruning resets), even on the same dataset. Since exhaustively enumerating these settings is beyond the scope of this survey, we link the original codebases and configurations in our \href{https://github.com/guikunchen/3DGS-Benchmarks}{benchmark repository} to support reproducibility. Likewise, memory and storage footprint are important practical factors but are highly benchmark-dependent, varying with the dataset/scene (\eg, scale, duration, texture complexity, and number of views), the target quality, and implementation choices. Consequently, GS methods may exhibit either smaller or larger VRAM/storage than NeRF baselines depending on the dataset and the chosen capacity. We therefore provide the same \href{https://github.com/guikunchen/3DGS-Benchmarks}{repository} to help readers reproduce and inspect footprint-quality trade-offs under specific datasets/settings of interest.

\subsection{Performance Benchmarking: Localization}
\label{sec:perm_local}
The localization task in SLAM involves determining the precise position and orientation of a robot or device.

\noindent$\bullet$~\textbf{Dataset:}
Replica~\cite{straub2019replica} dataset is a collection of 18 highly detailed 3D indoor scenes. These scenes are not only visually realistic but also offer data including dense meshes, high-quality textures, and detailed semantic information for each element. Following~\cite{sucar2021imap}, three sequences about rooms and five sequences about offices are used for the evaluation.

\noindent$\bullet$~\textbf{Benchmarking Algorithms:}
For performance comparison, we involve four recent 3D GS based algorithms~\cite{yugay2023gaussian,matsuki2023gaussian,keetha2023splatam,yan2023gs} and six typical SLAM methods~\cite{zhu2022nice,sandstrom2023point}.

\noindent$\bullet$~\textbf{Evaluation Metric:}
The root mean square error (RMSE) of the absolute trajectory error (ATE) is a commonly used metric in evaluating SLAM systems~\cite{sturm2012benchmark}, which measures the root mean square of the Euclidean distances between the estimated and true positions over the entire trajectory.

\noindent$\bullet$~\textbf{Result:}
As shown in Table~\ref{tbl:quanti_local}, the recent 3D Gaussians based localization algorithms have a clear advantage over existing NeRF based dense visual SLAM. For example, SplaTAM~\cite{keetha2023splatam} achieves a trajectory error improvement of $\sim$\textbf{50}\%, decreasing it from 0.52cm to \textbf{0.36cm} compared to the previous state-of-the-art (SOTA)~\cite{sandstrom2023point}. We attribute this to the dense and accurate 3D Gaussians reconstructed for scenes, which can handle the noise of real sensors. This reveals that effective scene representations can improve the accuracy of localization tasks.

\subsection{Performance Benchmarking: Static Scenes}
\label{sec:perm_render_static}
Rendering focuses on transforming computer-readable information (\eg, 3D objects in the scene) to pixel-based images. This section focuses on evaluating the quality of rendering results in static scenes.

\noindent$\bullet$~\textbf{Dataset:}
Replica~\cite{straub2019replica} (Sec.~\ref{sec:perm_local}) is used for comparison. The testing views are the same as \cite{sucar2021imap}.

\noindent$\bullet$~\textbf{Benchmarking Algorithms:}
For performance comparison, we involve four recent papers which introduce 3D Gaussians into their systems~\cite{yugay2023gaussian,matsuki2023gaussian,keetha2023splatam,yan2023gs}, as well as three dense SLAM methods~\cite{zhu2022nice,sandstrom2023point}.

\noindent$\bullet$~\textbf{Evaluation Metric:}
Peak signal-to-noise ratio (PSNR), structural similarity (SSIM)~\cite{wang2004image}, and learned perceptual image patch similarity (LPIPS)~\cite{zhang2018unreasonable} are used for measuring RGB rendering performance.

\noindent$\bullet$~\textbf{Result:}
Table~\ref{tbl:quanti_render_static} shows that 3D Gaussians based systems generally outperform the three dense SLAM competitors. For example, Gaussian-SLAM~\cite{yugay2023gaussian} establishes new SOTA and outperforms previous methods by a large margin. Compared to Point-SLAM~\cite{sandstrom2023point}, GSSLAM~\cite{matsuki2023gaussian} is about \textbf{578} times faster in achieving very competitive accuracy. In contrast to previous method~\cite{sandstrom2023point} that relies on depth information, such as depth-guided ray sampling, for synthesizing novel views, 3D GS based system eliminates this need, allowing for high-fidelity rendering for any views.

\subsection{Performance Benchmarking: Dynamic Scenes}
\label{sec:perm_render_dynamic}

This section focuses on evaluating the rendering quality in dynamic scenes.

\noindent$\bullet$~\textbf{Dataset:}
D-NeRF~\cite{pumarola2021d} dataset includes videos with 50 to 200 frames each, captured from unique viewpoints. It features synthetic, animated objects in complex scenes, with non-Lambertian materials. The dataset provides 50 to 200 training images and 20 test images per scene. The testing views are the same as the original paper~\cite{pumarola2021d}.

\noindent$\bullet$~\textbf{Benchmarking Algorithms:}
For performance comparison, we involve five recent papers that model dynamic scenes with 3D GS~\cite{yu2023cogs,wu20234d,lu20243d,yang2023deformable,yang2023real}, as well as six NeRF based approaches~\cite{pumarola2021d,guo2023forward,wang2023masked}.

\noindent$\bullet$~\textbf{Evaluation Metric:}
The same metrics as in Sec.~\ref{sec:perm_render_static}, \ie, PSNR, SSIM~\cite{wang2004image}, and LPIPS~\cite{zhang2018unreasonable}, are used for evaluation.

\noindent$\bullet$~\textbf{Result:}
From Table~\ref{tbl:quanti_render_dynamic} we can observe that 3D GS based methods outperform existing SOTAs by a clear margin. The static version of 3D GS~\cite{kerbl20233d} fails to reconstruct dynamic scenes, resulting in a sharp drop in performance. By modeling the dynamics, D-3DGS~\cite{yang2023deformable} outperforms the SOTA method, FFDNeRF~\cite{guo2023forward}, by \textbf{6.83}dB in terms of PSNR. These results indicate the effectiveness of introducing additional properties or structured information to model the deformation of Gaussians so as to model the scene dynamics.

\subsection{Performance Benchmarking: Human Avatar}
\label{sec:perm_avatar}

Human avatar modeling aims to create the model of human avatars from a given multi-view video.

\noindent$\bullet$~\textbf{Dataset:}
ZJU-MoCap~\cite{peng2021neural} is a prevalent benchmark in human modeling from videos, captured with 23 synchronized cameras at a 1024$\times$1024 resolution. Six subjects (\ie, 377, 386, 387, 392, 393, and 394) are used for evaluation~\cite{weng2022humannerf}. The same testing views following~\cite{geng2023learning} are adopted.

\noindent$\bullet$~\textbf{Benchmarking Algorithms:}
For performance comparison, we involve three recent papers which model human avatar with 3D GS~\cite{lei2023gart,hu2023gauhuman,qian20233dgs}, as well as six human rendering approaches~\cite{peng2021neural,weng2022humannerf,geng2023learning}.

\noindent$\bullet$~\textbf{Evaluation Metric:}
PSNR, SSIM~\cite{wang2004image}, and LPIPS*~\cite{zhang2018unreasonable} are used for measuring RGB rendering performance.

\noindent$\bullet$~\textbf{Result:}
Table~\ref{tbl:quanti_avatar} presents the numerical results of top-leading solutions in human avatar modeling. We observe that introducing 3D GS into the framework leads to consistent performance improvements in both rendering quality and speed. For instance, GART~\cite{lei2023gart} outperforms current SOTA, Instant-NVR~\cite{geng2023learning}, by \textbf{1.21}dB in terms of PSNR. Considering the enhanced fidelity, inference speed and editability, 3D GS based avatar modeling may revolutionize the field of 3D animation, interactive gaming, \etc. 

\subsection{Performance Benchmarking: Surgical Scenes}
\label{sec:perm_surgical}
3D reconstruction from endoscopic video is critical to robotic-assisted minimally invasive surgery.

\noindent$\bullet$~\textbf{Dataset:}
EndoNeRF~\cite{wang2022neural} dataset presents a specialized collection of stereo camera captures, comprising two samples of in-vivo prostatectomy. It is tailored to represent real-world surgical complexities, including challenging scenes with tool occlusion and pronounced non-rigid deformation. The same testing views as in~\cite{zha2023endosurf} are used.
   
\noindent$\bullet$~\textbf{Benchmarking Algorithms:}
For performance comparison, we involve three works which reconstruct dynamic 3D endoscopic scenes with GS~\cite{huang2024endo,liu2024endogaussian,zhao2024hfgs}, as well as three NeRF-based surgical reconstruction approaches~\cite{wang2022neural,zha2023endosurf}.

\noindent$\bullet$~\textbf{Evaluation Metric:}
PSNR, SSIM~\cite{wang2004image}, and LPIPS~\cite{zhang2018unreasonable} are adopted for evaluation.

\noindent$\bullet$~\textbf{Result:}
Table~\ref{tbl:quanti_surgical} shows that introducing the explicit representation of 3D Gaussians leads to several significant improvements. For instance, EndoGaussian~\cite{liu2024endogaussian} outperforms a strong baseline, EndoSurf~\cite{zha2023endosurf}, among all metrics. In particular, EndoGaussian demonstrates an approximate 200-fold increase in speed while consuming just 10\% of the GPU resources, a gap that mainly stems from the GS pipeline and lightweight dynamic modeling rather than unusually heavyweight NeRF baselines. These impressive results attest to the efficiency of GS-based methods, which not only expedite processing but also minimize GPU load, thus easing the demands on hardware. Such attributes are significant for surgical application deployment, where optimized resource usage can be a key determinant of practical utility.

\section{Future Research Directions}
\label{sec:direction}

\textbf{Recent Trends and The Big Picture.}
Recent work on GS has pushed it from a fast alternative to NeRFs towards a general representation for 3D world modeling. Early work showed that explicit 3D Gaussians can be optimized to capture complex static and dynamic scenes with high fidelity, inspiring follow-up research on more compact, generalizable, and semantics-aware splat-based models~\cite{lee2023compact,liu2024fast,gao20253d}. As an explicit scene representation, GS has been applied across a wide range of vision, graphics, and robotics applications~\cite{tang2023dreamgaussian,abou2023physically,feng2025gaussian,abou2024physically}.
A major recent trend is the shift from per-scene optimization to feed-forward Gaussian prediction. Early pioneering works showed that 3D Gaussians can be inferred directly from sparse views using lightweight, pixel-level feature extractors (\eg, U-Nets)~\cite{charatan2023pixelsplat,szymanowicz2023splatter}. To improve scalability from object-centric reconstructions to complex scenes, more recent approaches have adopted Transformer-based models~\cite{GSLRM,xu2024grm}. This architectural shift has naturally driven efforts to process increasingly large input sets; by innovating on sequence processing -- such as integrating hybrid attention mechanisms~\cite{LongLRM,ziwen2025long} or test-time training~\cite{LaCT,wang2026tttlrm} -- current models can now ingest massive multi-view contexts for wide-coverage, scene-level modeling. Simultaneously, this feed-forward paradigm is rapidly expanding its functional scope: recent frameworks can now jointly infer camera parameters from uncalibrated image collections~\cite{AnySplat}, extract robust geometric priors to stabilize downstream prediction tasks~\cite{wang2025vggt}, and extend spatial prediction into the temporal domain for 4D dynamic generation~\cite{L4GM,ma20254d}.
Together, these trends point towards Gaussian-based world models that support not only high-quality rendering and reconstruction but also interactive perception, reasoning, and action, with demonstrations in robotics and embodied AI. Concrete discussions on some important directions are given in the following.

\noindent$\bullet$~\textbf{Physics- and Semantics-aware Scene Representation.}
As a new, explicit scene representation technique, 3D Gaussian offers transformative potential beyond merely enhancing novel-view synthesis. It has the potential to pave the way for simultaneous advancements in scene reconstruction and understanding by devising physics- and semantics-aware 3D GS systems. While significant progress has been made in physics (Sec.~\ref{sec:app_physics}) and semantics~\cite{wang2024query,qu2024goi,ji2024fastlgs,liao2024clip,choi2024click,ji2024segment} individually, there remains considerable untapped potential in their synergistic integration. This is poised to revolutionize a range of fields and downstream applications. For instance, incorporating prior knowledge such as the general shape of objects can reduce the need for extensive training viewpoints~\cite{szymanowicz2023splatter,charatan2023pixelsplat} while improving geometry/surface reconstruction~\cite{guedon2023sugar,li2024geogaussian}. A critical metric for assessing scene representation is the quality of its generated scenes, which encompasses challenges in geometry, texture, and lighting fidelity~\cite{huang2023sc,chen2023gaussianeditor,gao2023relightable}. By merging physical principles and semantic information within the 3D GS framework, one can expect that the quality will be enhanced, thereby facilitating dynamics modeling~\cite{xie2023physgaussian,jiang2024vr}, editing~\cite{cen2023segment,zhou2023feature}, generation~\cite{tang2023dreamgaussian,yi2023gaussiandreamer}, and beyond. In a nutshell, pursuing this advanced and versatile scene representation opens up new possibilities for innovation in computational creativity and practical applications across diverse domains.

\noindent$\bullet$~\textbf{Learning Physical Priors from Large-scale Data.}
As we explore the potential of physics- and semantics-aware scene representations, leveraging large-scale datasets to learn generalizable, physical priors emerges as a promising direction. The goal is to model the inherent physical properties and dynamics embedded within real-world data, transforming them into actionable insights that can be applied across various domains such as robotics and visual effects. Establishing a learning framework for extracting these generalizable priors enables the application of these insights to specific tasks in a few-shot manner. For instance, it allows for rapid adaptation to new objects and environments with minimal data input. Furthermore, integrating physical priors can enhance not only the accuracy and quality of generated scenes but also their interactive and dynamic qualities. This is particularly valuable in AR/VR environments, where users interact with virtual objects that behave in ways consistent with their real-world counterparts. However, the existing body of work on capturing and distilling physics-based knowledge from extensive 2D and 3D datasets remains sparse. Notable efforts in related area include the continuum mechanics based GS systems (Sec.~\ref{sec:app_physics}), and the generalizable Gaussian representation based on multi-view stereo~\cite{liu2024fast}. Further exploration on real2sim and sim2real might offer viable routes for advancements in this field.

\noindent$\bullet$~\textbf{Modeling Internal Structures of Objects with 3D GS.}
Despite the ability of 3D GS to produce highly photorealistic renderings, modeling internal structures of objects (\eg, for a scanned object in computed tomography) within the current GS framework presents a notable challenge. Due to the splatting and density control process, the current representation of 3D Gaussian is unorganized and cannot align well with the object's actual internal structures. Moreover, there is a strong preference in various applications to depict objects as volumes (\eg, computed tomography). However, the disordered nature of 3D GS makes volume modeling particularly difficult. Li \etal \cite{li2023sparse} used 3D Gaussians with density control as the basis for the volumetric representation and did not involve the splatting process. X-Gaussian~\cite{cai2024radiative} involves the splatting process for fast training and inference but cannot generate volumetric representation. Using 3D GS to model the internal structures of objects remains unanswered and deserves further exploration.

\noindent$\bullet$~\textbf{3D GS for Simulation in Autonomous Driving and beyond.}
Collecting real-world datasets for autonomous driving is both expensive and logistically challenging, yet crucial for training effective image-based perception systems. To mitigate these issues, simulation emerges as a cost-effective alternative, enabling the generation of synthetic datasets across diverse environments. However, the development of simulators capable of producing photorealistic and diverse synthetic data is fraught with challenges. These include achieving a high level of quality, accommodating various control methods, and accurately simulating a range of lighting conditions. While early efforts~\cite{zhou2023drivinggaussian,yan2024street,zhou2024hugs} in reconstructing urban/street scenes with 3D GS have been encouraging, they are just the tip of the iceberg in terms of the full capabilities. There remain numerous critical aspects to be explored, such as the integration of user-defined object models, the modeling of physics-aware scene changes (\eg, the rotation of vehicle wheels), and the enhancement of controllability and quality (\eg, in varying lighting conditions). Mastery of these capabilities would not only advance autonomous systems but also redefine computational understanding of physical spaces --- a leap with implications for world models, spatial intelligence, embodied AI, and beyond.

\noindent$\bullet$~\textbf{Empowering 3D GS with More Possibilities.}
Despite the significant potential of 3D GS, the full scope of applications for 3D GS remains largely untapped. A promising avenue for exploration involves augmenting 3D Gaussians with additional attributes (\eg, linguistic and spatiotemporal properties as mentioned in Sec. \ref{sec:e3dgs_property}) and introducing structured information (e.g., spatial MLPs and grids as mentioned in Sec. \ref{sec:e3dgs_struct}), tailored for specific applications. Moreover, recent studies have begun to unveil the capability of 3D GS in several domains, \eg, point cloud registration~\cite{chang2024gaussreg}, image representation and compression~\cite{zhang2024gaussianimage}, and fluid synthesis~\cite{feng2024gaussian}. These findings highlight a significant opportunity for interdisciplinary scholars to explore 3D GS further.

\section{Conclusions}
\label{sec:conclusions}

To the best of our knowledge, this survey presents the first comprehensive overview of 3D GS, a groundbreaking technique revolutionizing explicit radiance fields, computer graphics, and computer vision. It delineates the paradigm shift from traditional NeRF based methods, spotlighting the advantages of 3D GS in real-time rendering and enhanced editability. Our in-depth analysis and extensive quantitative studies demonstrate the superiority of 3D GS in practical applications, particularly those highly sensitive to latency. We offer insights into principles, prospective research directions, and the unresolved challenges within this domain. Overall, 3D GS stands as a transformative technology, poised to significantly influence future advancements in 3D reconstruction and representation. This survey is intended to serve as a foundational resource, propelling further exploration and progress in this rapidly evolving field.

\bibliographystyle{ACM-Reference-Format}
\balance
\bibliography{egbib}
\clearpage

\section*{Supplementary Material}

\setcounter{table}{0}
\renewcommand{\thetable}{S\arabic{table}}

\setcounter{figure}{0}
\renewcommand{\thefigure}{S\arabic{figure}}

Due to the page limit of the main manuscript, we provide additional figures and tables in this supplementary material.
Specifically, we include \textbf{i}) a visualization of the 3D GS growth process; \textbf{ii}) a visual comparison between NeRFs and 3D GS;  \textbf{iii}) a demonstration of tile-based parallel rendering; \textbf{iv}) a consolidated summary table of the commonly used datasets; and \textbf{v}) detailed quantitative performance-comparison tables. These figures and tables are referenced from the main text and are intended to improve completeness without interrupting the main narrative.

\begin{figure}[h]
    \centering
    \includegraphics[width=0.5\textwidth]{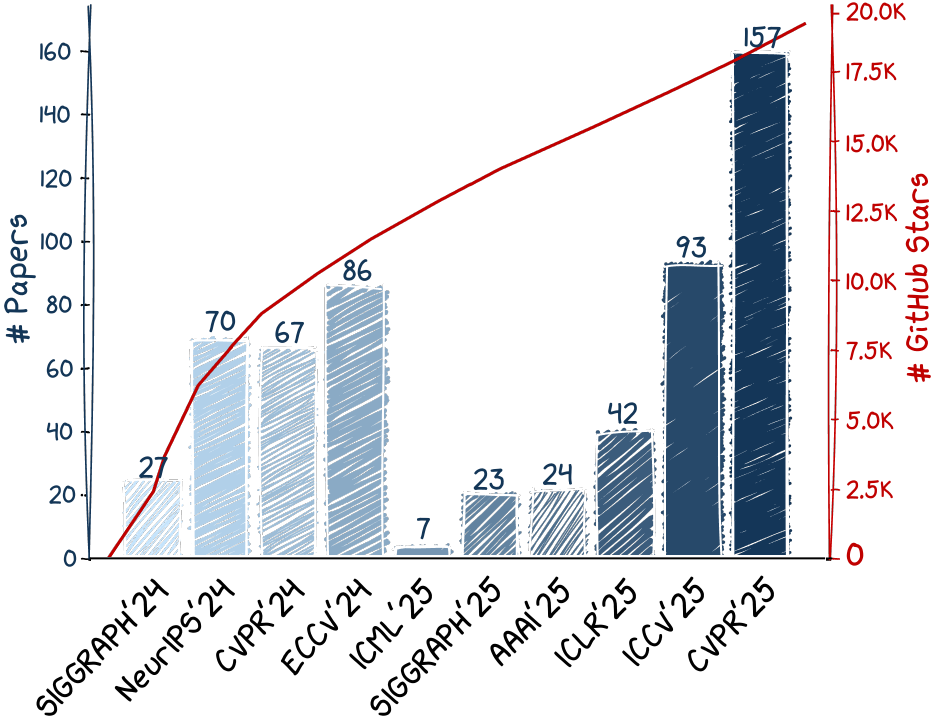}
    \caption{The number of published papers and GitHub stars on 3D GS. Statistics sourced from \href{https://github.com/Awesome3DGS/3D-Gaussian-Splatting-Papers}{\# Papers} and \href{https://star-history.com/\#graphdeco-inria/gaussian-splatting\&Date}{\# GitHub Stars}.}
    \label{fig_num_paper}
\end{figure}

\begin{figure}[h]
    \centering
    \includegraphics[width=0.5\textwidth]{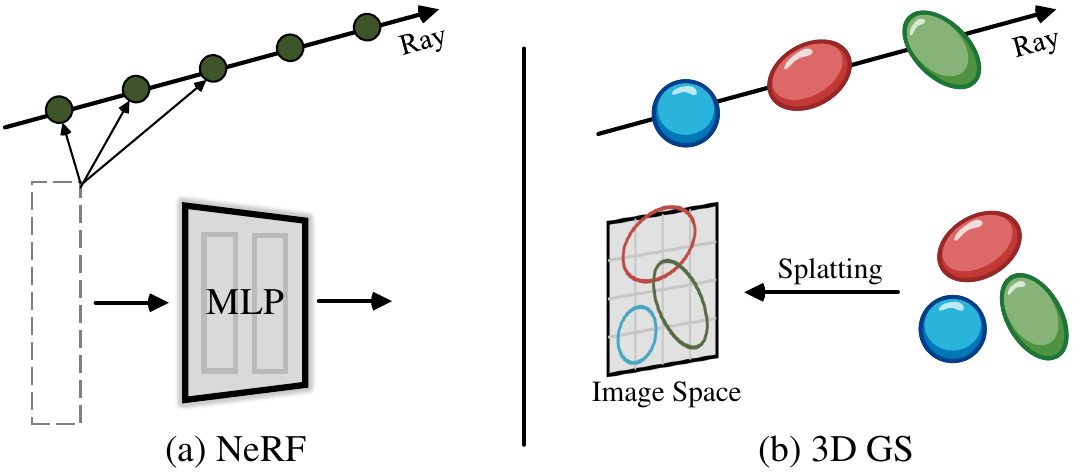}
    \put(-207.5, 20){\footnotesize\rotatebox{90}{$x, y, z, \theta, \phi$}}
    \put(-135, 33){\footnotesize{$c ~\&~ \sigma$}}
    \caption{NeRFs \vs 3D GS. (a) NeRF samples along the ray and then queries the MLP to obtain colors and densities, which can be seen as a \emph{backward} mapping (ray tracing). (b) 3D GS projects all 3D Gaussians into the image space (\ie, splatting) and then performs parallel rendering, which can be viewed as a \emph{forward} mapping (rasterization).}
    \label{fig:fig_nerf_3dgs}
\end{figure}



\begin{figure}[h]
    \centering
    \includegraphics[width=0.5\textwidth]{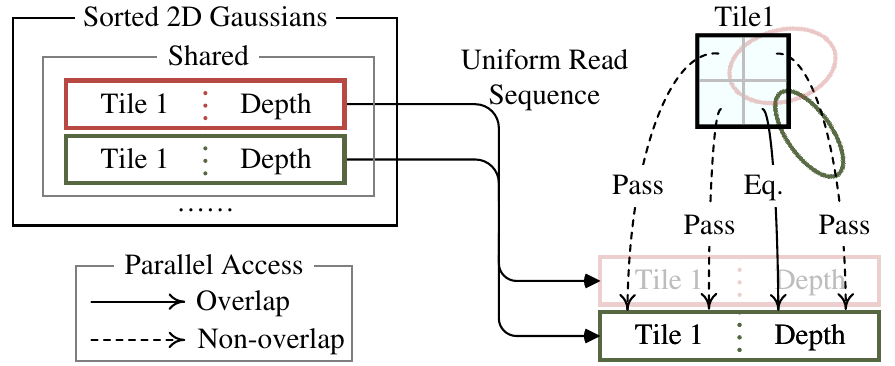}
    \put(-25, 41){\footnotesize{\ref{eq:alpha_compo}}} 
    \caption{An illustration of the tile based parallel (at the pixel-level) rendering. All the pixels within a tile (Tile1 here) access the same ordered Gaussian list stored in a shared memory for rendering. As the system processes each Gaussian sequentially, every pixel in the tile evaluates the Gaussian's contribution according to the distance (i.e., the $\exp$ term in Eq.~\ref{eq:gaussian_opacity}). Therefore, the rendering for a tile can be completed by iterating through the list of Gaussians just once. The computation for the \textcolor{3dgc1}{red} Gaussian follows a similar way and is omitted here for simplicity.}
    \label{fig:fig_tile_rendering}
\end{figure}



\begin{table}[h]
    \centering
    \captionsetup{font=small}
    \caption{Summary of major 3DGS research directions reviewed in Section~\ref{sec:e3dgs}. For each direction, representative works are listed in chronological order, providing a rough timeline of progress.}
    \label{tbl:direction_summary}
    \vspace{-1em}
    \footnotesize
    \resizebox{\linewidth}{!}{%
    \begin{tabular}{p{0.1\linewidth} || p{0.20\linewidth} | p{0.13\linewidth} | p{0.23\linewidth} | p{0.26\linewidth}}
        \hline\thickhline
        \rowcolor{mygray}
    \textbf{Direction} & \textbf{Key ideas} & \textbf{Representative works} & \textbf{Advantages} & \textbf{Limitations / challenges} \\
    \hline\hline
    \S\ref{sec:e3dgs_de3dgs} 3DGS for Sparse Input 
     & Handle few-view / sparse inputs via regularization or learned priors
     & \cite{xiong2023sparsegs,zhu2023fsgs,charatan2023pixelsplat,szymanowicz2023splatter,li2024dngaussian,swann2024touch,chen2024mvsplat,wewer2024latentsplat,xu2024grm,shen2024gamba,zhang2024cor}
     & Supports limited viewpoints; enables single-image or few-shot NVS; leverages strong 3D priors
     & Trade-off between geometry quality and speed; sensitive to prior/regularization choice; dynamic scenes remain difficult \\
    \hline
    \S\ref{sec:e3dgs_me3dgs}  Memory-efficient 3DGS 
     & Reduce Gaussian count and/or compress Gaussian attributes for compact storage
     & \cite{fan2023lightgaussian,navaneet2023compact3d,lee2023compact,morgenstern2023compact,zhang2024gaussianimage,niedermayr2024compressed,chen2024hac,papantonakis2024reducing,fang2024mini}
     & Lower memory footprint; scalable to large scenes and datasets; facilitates deployment and streaming
     & Quality drop at high compression; training-time memory is still high; designing reusable codebooks is open \\
    \hline
    \S\ref{sec:e3dgs_photo}  Photorealistic 3D GS 
     & Improve aliasing, reflection, blur, and shading to boost visual fidelity
     & \cite{yu2023mip,yan2023multi,jiang2023gaussianshader,lee2024deblurring,bolanos2024gaussian,radl2024stopthepop,yang2024spec,zhao2024bad,dahmani2024swag,li2024geogaussian,liang2024analytic,seiskari2024gaussian,song2024sa}
     & Sharper details; more realistic materials and lighting; increased robustness on real captures
     & Pipelines become more complex; many methods target specific artifacts; may increase training and rendering cost \\
    \hline
    \S\ref{sec:e3dgs_optim}  Improved Optimization Algorithms 
     & Better optimization and regularization pipelines for more stable training
     & \cite{fu2023colmap,lu2023scaffold,jung2024relaxing,li2024geogaussian,yu2024gsdf,zhang2024fregs,huang2024gs++,li2024loopgaussian}
     & Faster convergence; fewer floaters/artifacts; more robust reconstruction in challenging scenes
     & Still per-scene optimization; hyperparameter and schedule tuning; limited work on meta-learning or few-shot GS \\
     \hline
     \S\ref{sec:e3dgs_property}  3D Gauss-ians with More Properties 
     & Attach semantic or spatiotemporal features to Gaussians for richer scene understanding
     & \cite{shi2023language,qin2023langsplat,zhou2023feature,ye2023gaussian,cen2023segment,yang2023real,zuo2024fmgs}
     & Enables open-vocabulary querying, semantic editing, spatiotemporal modeling, \textit{etc}.
     & Higher memory and computation; depends on large-scale supervision; Evaluation protocols are still emerging \\
     \hline
     \S\ref{sec:e3dgs_struct}  Hybrid Re-presentations 
     & Combine Gaussians with structured modules (MLPs, deformation fields) for specific tasks
     & \cite{yang2023deformable,wu20234d,xu2023gaussian,saroha2024gaussian}
     & Adds structure and controllability; effective for avatars, dynamic scenes, stylization, \textit{etc}.
     & More complex architectures; less generic; partially loses the simplicity of pure GS \\
     \hline
     \S\ref{sec:e3dgs_new_rendering} New Rendering Algorithms 
     & Devise ray tracing / physically-based and improved rasterization for Gaussians
     & \cite{moenne20243d,mai2024ever,condor2025don}
     & Better handling of visibility, secondary rays, and complex cameras; improved physical correctness
     & Higher computational cost; integration with real-time pipelines is still in early stages; APIs and tooling are immature \\
    \hline
    \end{tabular}
    }
\end{table}

\begin{table}[h]
    \centering
    \captionsetup{font=small}
    \caption{Comparison of regularization-based and generalizability-based 3DGS methods for sparse input.}
    \vspace{-1em}
    \small
    \resizebox{0.95\textwidth}{!}{
      \setlength\tabcolsep{13pt}{}
      \begin{tabular}{c||c|c}
        \hline\thickhline
        \rowcolor{mygray}
        Dimension & Regularization-based & Generalizability-based \\
        \hline\hline
        Inference speed  & Slow; per-scene optimization  & Fast; feed-forward \\
        Reconstruction quality  & High detail; strong geometry  & Lower accuracy; priors may hallucinate \\
        Views required at test time  & Typically multi-view  & Few views; even single image \\
        Robustness to unseen regions  & Limited; overfits observed views  & Depends on prior; plausible but inexact \\
        Typical use cases  & High-fidelity reconstruction  & Real-time / interactive NVS \\
        Representative works  & \cite{li2024dngaussian,zhu2023fsgs,chen2024mvsplat,zhang2024cor} & \cite{charatan2023pixelsplat,szymanowicz2023splatter,xu2024grm,szymanowicz2024flash3d} \\
    \hline
    \end{tabular}
    }
    \label{tbl:sparse_tradeoff}
    \vspace{-1em}
\end{table}

\begin{table}[b]
    \centering
    \caption{ Collection of representative datasets for 3D GS. Here PC represents point clouds.}
    \small
    \resizebox{0.72\textwidth}{!}{
      \setlength\tabcolsep{6pt}{}
      \begin{tabular}{z{82}y{18}||cc||c}
        \hline\thickhline
        \rowcolor{mygray}
        \multicolumn{2}{c||}{Name} & Type & \# Sample & Area and Task \\
        \hline\hline
        
        \href{https://www.tanksandtemples.org/}{Tanks\&Temples}~\cite{knapitsch2017tanks}\!\!&\!\!\!\pub{TOG17}\! & RGB & 14 & \multirow{7}{*}{\shortstack{Novel View Synthesis}} \\
        \href{https://google.github.io/realestate10k/}{RealEstate10K}~\cite{zhou2018stereo}\!\!&\!\!\!\pub{TOG18}\! & RGB & 1,000 & \\
        \href{http://visual.cs.ucl.ac.uk/pubs/deepblending/}{DeepBlending}~\cite{hedman2018deep}\!\!&\!\!\!\pub{TOG18}\! & RGB & 19 & \\
        \href{https://bmild.github.io/llff/}{LLFF}~\cite{mildenhall2019local}\!\!&\!\!\!\pub{TOG19}\! & RGB & 8 & \\
        \href{https://www.matthewtancik.com/nerf}{NeRF}~\cite{mildenhall2020nerf}\!\!&\!\!\!\pub{ECCV20}\! & RGB & 8 & \\
        \href{https://infinite-nature.github.io/}{ACID}~\cite{liu2021infinite}\!\!&\!\!\!\pub{ICCV21}\! & RGB & 700+ & \\
        \href{https://jonbarron.info/mipnerf360/}{Mip-NeRF 360}~\cite{barron2022mip}\!\!&\!\!\!\pub{CVPR22}\! & RGB & 9 & \\
        \hline

        \href{https://cvg.cit.tum.de/data/datasets/rgbd-dataset}{TUM RGB-D}~\cite{sturm2012benchmark}\!\!&\!\!\!\pub{IROS12}\! & RGB-D & 39 & \multirow{9}{*}{Robotics} \\
        \href{http://www.cvlibs.net/datasets/kitti/}{KITTI}~\cite{geiger2012we}\!\!&\!\!\!\pub{CVPR12}\! & \makecell{RGB-D\&PC} & 11 & \\
        \href{http://www.scan-net.org/}{ScanNet}~\cite{dai2017scannet}\!\!&\!\!\!\pub{CVPR17}\! & RGB-D & 1,513 & \\
        \href{https://github.com/facebookresearch/Replica-Dataset}{Replica}~\cite{straub2019replica}\!\!&\!\!\!\pub{arXiv19}\! & RGB-D & 18 &  \\
        \href{http://www.scan-net.org/}{Waymo}~\cite{sun2020scalability}\!\!&\!\!\!\pub{CVPR20}\! & \makecell{RGB-D\&PC} & 1,150 & \\
        \href{http://www.scan-net.org/}{nuScenes}~\cite{caesar2020nuscenes}\!\!&\!\!\!\pub{CVPR20}\! & \makecell{RGB-D\&PC} & 1,000 & \\
        \href{https://github.com/stepjam/RLBench}{RLBench}~\cite{james2020rlbench}\!\!&\!\!\!\pub{RA-L20}\! & RGB & 100 & \\
        \href{https://robomimic.github.io/}{Robomimic}~\cite{mandlekarmatters}\!\!&\!\!\!\pub{CoRL22}\! & RGB & 800 & \\
        \hline

        \href{https://github.com/albertpumarola/D-NeRF?tab=readme-ov-file\#download-dataset}{D-NeRF}~\cite{pumarola2021d}\!\!&\!\!\!\pub{CVPR21}\! & RGB & 8 & \multirow{3}{*}{\shortstack{Dynamic Scene  Reconstruction}} \\
        \href{https://github.com/google/hypernerf/releases/tag/v0.1}{HyperNeRF}~\cite{park2021hypernerf}\!\!&\!\!\!\pub{TOG21}\! & RGB & 6 & \\
        \href{https://github.com/JokerYan/NeRF-DS?tab=readme-ov-file\#data}{NeRF-DS}~\cite{yan2023nerf}\!\!&\!\!\!\pub{CVPR23}\! & RGB & 8 & \\
        \hline

        \href{https://github.com/kacperkan/conerf}{CoNeRF}~\cite{kania2022conerf}\!\!&\!\!\!\pub{CVPR22}\! & RGB & 7 & \multirow{5}{*}{\shortstack{Generation and Editing}} \\
        \href{https://spinnerf3d.github.io/}{SPIn-NeRF}~\cite{mirzaei2023spin}\!\!&\!\!\!\pub{CVPR23}\! & RGB & 10 & \\
        \href{https://github.com/DSaurus/Tensor4D}{Tensor4D}~\cite{shao2023tensor4d}\!\!&\!\!\!\pub{CVPR23}\! & RGB & 4 & \\
        \href{https://opendatalab.com/OpenXD-OmniObject3D-New/download}{OmniObject3D}~\cite{wu2023omniobject3d}\!\!&\!\!\!\pub{CVPR23}\! & 3D Object & 6,000 & \\
        \href{https://objaverse.allenai.org/objaverse-1.0}{Objaverse}~\cite{deitke2023objaverse}\!\!&\!\!\!\pub{CVPR23}\! & 3D Object & 800K+ & \\
        \hline

        \href{https://graphics.tu-bs.de/people-snapshot}{ People-Snapshot}~\cite{alldieck2018video}\!\!&\!\!\!\pub{CVPR18}\! & RGB & 24 & \multirow{7}{*}{\shortstack{Avatar}} \\
        \href{https://voca.is.tue.mpg.de/}{VOCASET}~\cite{cudeiro2019capture}\!\!&\!\!\!\pub{CVPR19}\! & RGB & 12 & \\
        \href{https://github.com/ZhengZerong/DeepHuman/tree/master/THUmanDataset}{THUman}~\cite{zheng2019deephuman}\!\!&\!\!\!\pub{ICCV19}\! & RGB & 200 & \\
        \href{https://github.com/ytrock/THuman2.0-Dataset}{THUman2.0}~\cite{yu2021function4d}\!\!&\!\!\!\pub{CVPR21}\! & RGB-D & 500 & \\
        \href{https://github.com/zju3dv/neuralbody/blob/master/INSTALL.md\#zju-mocap-dataset}{ZJU-Mocap}~\cite{peng2021neural}\!\!&\!\!\!\pub{CVPR21}\! & RGB & 9 &  \\
        \href{https://crisalixsa.github.io/h3d-net/}{H3DS}~\cite{ramon2021h3d}\!\!&\!\!\!\pub{ICCV21}\! & RGB & 23 & \\
        \href{https://github.com/fwbx529/THuman3.0-Dataset}{THUman3.0}~\cite{su2022deepcloth}\!\!&\!\!\!\pub{TPAMI22}\! & 3D Scan & 20 & \\
        
        \hline

        \href{https://endovissub2019-scared.grand-challenge.org/}{SCARED}~\cite{allan2021stereo}\!\!&\!\!\!\pub{MICCAI19}\! & RGB-D & 9 & \multirow{3}{*}{\shortstack{Medical Systems}} \\
        \href{https://github.com/med-air/EndoNeRF}{EndoNeRF}~\cite{wang2022neural}\!\!&\!\!\!\pub{MICCAI22}\! & RGB & 2 &  \\
        \href{https://github.com/caiyuanhao1998/SAX-NeRF?tab=readme-ov-file\#2-prepare-dataset}{X3D}~\cite{cai2024structure}\!\!&\!\!\!\pub{CVPR24}\! & X-ray & 15 & \\
        \hline

        \href{https://github.com/city-super/BungeeNeRF}{CityNeRF}~\cite{xiangli2022bungeenerf}\!\!&\!\!\!\pub{ECCV22}\! & RGB & 12 & \multirow{4}{*}{\shortstack{Large-scale Reconstrction}} \\
        \href{https://waymo.com/research/block-nerf}{Waymo Block-NeRF}~\cite{tancik2022block}\!\!&\!\!\!\pub{CVPR22}\! & RGB\&PC & 1 &  \\
        \href{https://vcc.tech/UrbanBIS}{UrbanBIS}~\cite{yang2023urbanbis}\!\!&\!\!\!\pub{SIGGR23}\! & RGB\&PC & 6 &  \\
        \href{https://saliteta.github.io/CUHKSZ_SMBU/}{GauU-Scene}~\cite{xiong2024gauu}\!\!&\!\!\!\pub{arXiv24}\! & RGB\&PC & 1 &  \\
         
        \hline
      \end{tabular}
    }
    \label{tbl:dataset}
\end{table}

\clearpage

\begin{table*}[t]
    \centering
    \caption{Comparison of \textbf{localization} methods on Replica~\cite{straub2019replica} (static scenes), in terms of absolute trajectory error (ATE, cm). (The three best scores are marked in \textcolor{red}{\textbf{red}}, \textcolor{blue}{\textbf{blue}}, and \textcolor{green}{\textbf{green}}, respectively. These notes also apply to the other tables.)}
    \small
    \resizebox{1.0\textwidth}{!}{
      \setlength\tabcolsep{7pt}{}
      \begin{tabular}{z{90}y{20}|c||cccccccc|c}
        \hline\thickhline
        \rowcolor{mygray}
        \multicolumn{2}{c|}{Method} & GS & Room0 & Room1 & Room2 & Office0 & Office1 & Office2 & Office3 & Office4 & Avarage \\
        \hline\hline
        iMAP~\cite{sucar2021imap}\!\!\!&\!\!\!\pub{ICCV21}\!&  & 3.12 & 2.54 & 2.31 & 1.69 & 1.03 & 3.99 & 4.05 & 1.93 & 2.58 \\
        Vox-Fusion~\cite{yang2022vox}\!\!\!&\!\!\!\pub{ISMAR22}\!&  & 1.37 & 4.70 & 1.47 & 8.48 & 2.04 & 2.58 & 1.11 & 2.94 & 3.09 \\
        NICE-SLAM~\cite{zhu2022nice}\!\!\!&\!\!\!\pub{CVPR22}\!&  & 0.97 & 1.31 & 1.07 & 0.88 & 1.00 & 1.06 & 1.10 & 1.13 & 1.06 \\
        ESLAM~\cite{johari2023eslam}\!\!\!&\!\!\!\pub{CVPR23}\!&  & 0.71 & 0.70 & 0.52 & 0.57 & 0.55 & 0.58 & 0.72 & \textcolor{green}{0.63} & 0.63 \\
        Point-SLAM~\cite{sandstrom2023point}\!\!\!&\!\!\!\pub{ICCV23}\!&  & 0.61 & \textcolor{blue}{0.41} & 0.37 & \textcolor{red}{0.38} & \textcolor{green}{0.48} & \textcolor{green}{0.54} & 0.69 & 0.72 & \textcolor{green}{0.52} \\
        Co-SLAM~\cite{wang2023co}\!\!\!&\!\!\!\pub{CVPR23}\!&  & 0.70 & 0.95 & 1.35 & 0.59 & 0.55 & 2.03 & 1.56 & 0.72 & 1.00 \\

        Gaussian-SLAM~\cite{yugay2023gaussian}\!\!\!&\!\!\!\pub{arXiv}\!& \checkmark & 3.35 & 8.74 & 3.13 & 1.11 & 0.81 & 0.78 & 1.08 & 7.21 & 3.27 \\
        GSSLAM~\cite{matsuki2023gaussian}\!\!\!&\!\!\!\pub{CVPR24}\!& \checkmark & \textcolor{blue}{0.47} & \textcolor{green}{0.43} & \textcolor{blue}{0.31} & 0.70 & 0.57 & \textcolor{blue}{0.31} & \textcolor{red}{0.31} & 3.20 & 0.79 \\
        GS-SLAM~\cite{yan2023gs}\!\!\!&\!\!\!\pub{CVPR24}\!& \checkmark & \textcolor{green}{0.48} & 0.53 & \textcolor{green}{0.33} & \textcolor{green}{0.52} & \textcolor{blue}{0.41} & 0.59 & \textcolor{green}{0.46} & \textcolor{green}{0.70} & \textcolor{blue}{0.50} \\
        SplaTAM~\cite{keetha2023splatam}\!\!\!&\!\!\!\pub{CVPR24}\!& \checkmark & \textcolor{red}{0.31} & \textcolor{red}{0.40} & \textcolor{red}{0.29} & \textcolor{blue}{0.47} & \textcolor{red}{0.27} & \textcolor{red}{0.29} & \textcolor{blue}{0.32} & \textcolor{red}{0.55} & \textcolor{red}{0.36} \\

        \hline
       \end{tabular}
    }
    \label{tbl:quanti_local}
    \vspace{-10pt} 
\end{table*}

         

\begin{table*}[t]
    \centering
    \caption{Comparison of \textbf{mapping} methods on Replica~\cite{straub2019replica} (static scenes), in terms of PSNR, SSIM, and LPIPS. FPS is measured using NVIDIA 4090~\cite{matsuki2023gaussian}.}
    \small
    \resizebox{0.97\textwidth}{!}{
      \setlength\tabcolsep{5pt}{}
      \begin{tabular}{z{80}y{20}|c||c|cccccccc|c|c}
         \hline\thickhline
         \rowcolor{mygray}
         \multicolumn{2}{c|}{Method} & GS & Metric & Room0 & Room1 & Room2 & Office0 & Office1 & Office2 & Office3 & Office4 & Avarage & FPS \\
         \hline\hline
         \multirow{3}{*}{NICE-SLAM~\cite{zhu2022nice}\!\!\!} & \multirow{3}{*}{\!\!\!\pub{CVPR22}\!} &  & PSNR$\uparrow$ & 22.12 & 22.47 & 24.52 & 29.07 & 30.34 & 19.66 & 22.23 & 24.94 & 24.42 & \multirow{3}{*}{0.54} \\
         &  &  & SSIM$\uparrow$ & 0.69 & 0.76 & 0.81 & 0.87 & 0.89 & 0.80 & 0.80 & 0.86 & 0.81 & \\
         &  &  & LPIPS$\downarrow$ & 0.33 & 0.27 & 0.21 & 0.23 & 0.18 & 0.23 & 0.21 & 0.20 & 0.23 & \\
         \hline

         \multirow{3}{*}{Vox-Fusion~\cite{yang2022vox}\!\!\!} & \multirow{3}{*}{\!\!\!\pub{ISMAR22}\!} &  & PSNR$\uparrow$ & 22.39 & 22.36 & 23.92 & 27.79 & 29.83 & 20.33 & 23.47 & 25.21 & 24.41 & \multirow{3}{*}{2.17} \\
         &  &  & SSIM$\uparrow$ & 0.68 & 0.75 & 0.80 & 0.86 & 0.88 & 0.79 & 0.80 & 0.85 & 0.80 & \\
         &  &  & LPIPS$\downarrow$ & 0.30 & 0.27 & 0.23 & 0.24 & 0.18 & 0.24 & 0.21 & 0.20 & 0.24 & \\
         \hline

         \multirow{3}{*}{Point-SLAM~\cite{sandstrom2023point}\!\!\!} & \multirow{3}{*}{\!\!\!\pub{ICCV23}\!} &  & PSNR$\uparrow$ & 32.40 & \textcolor{green}{34.08} & \textcolor{green}{35.50} & 38.26 & 39.16 & \textcolor{green}{33.99} & \textcolor{green}{33.48} & \textcolor{green}{33.49} & \textcolor{green}{35.17} & \multirow{3}{*}{1.33} \\
         &  &  & SSIM$\uparrow$ & \textcolor{green}{0.97} & \textcolor{blue}{0.98} & \textcolor{blue}{0.98} & \textcolor{green}{0.98} & \textcolor{green}{0.99} & 0.96 & 0.96 & \textcolor{blue}{0.98} & \textcolor{green}{0.97} & \\
         &  &  & LPIPS$\downarrow$ & 0.11 & 0.12 & 0.11 & 0.10 & 0.12 & 0.16 & 0.13 & 0.14 & 0.12 & \\
         \hline

         \multirow{3}{*}{SplaTAM~\cite{keetha2023splatam}\!\!\!} & \multirow{3}{*}{\!\!\!\pub{CVPR24}\!} & \multirow{3}{*}{\checkmark} & PSNR$\uparrow$ & \textcolor{green}{32.86} & 33.89 & 35.25 & 38.26 & 39.17 & 31.97 & 29.70 & 31.81 & 34.11 & \multirow{3}{*}{-} \\
         &  &  & SSIM$\uparrow$ & \textcolor{blue}{0.98} & 0.97 & \textcolor{green}{0.98} & 0.98 & 0.98 & \textcolor{green}{0.97} & 0.95 & 0.95 & 0.97 & \\
         &  &  & LPIPS$\downarrow$ & \textcolor{blue}{0.07} & 0.10 & \textcolor{green}{0.08} & 0.09 & 0.09 & 0.10 & 0.12 & 0.15 & 0.10 & \\
         \hline

         \multirow{3}{*}{GS-SLAM~\cite{yan2023gs}\!\!\!} & \multirow{3}{*}{\!\!\!\pub{CVPR24}\!} & \multirow{3}{*}{\checkmark} & PSNR$\uparrow$ & 31.56 & 32.86 & 32.59 & \textcolor{green}{38.70} & \textcolor{green}{41.17} & 32.36 & 32.03 & 32.92 & 34.27 & \multirow{3}{*}{-} \\
         &  &  & SSIM$\uparrow$ & 0.97 & \textcolor{green}{0.97} & 0.97 & \textcolor{blue}{0.99} & \textcolor{blue}{0.99} & \textcolor{blue}{0.98} & \textcolor{blue}{0.97} & \textcolor{green}{0.97} & \textcolor{blue}{0.97} & \\
         &  &  & LPIPS$\downarrow$ & 0.09 & \textcolor{red}{0.07} & 0.09 & \textcolor{blue}{0.05} & \textcolor{red}{0.03} & \textcolor{green}{0.09} & \textcolor{green}{0.11} & \textcolor{green}{0.11} & \textcolor{green}{0.08} & \\
         \hline
         
         \multirow{3}{*}{GSSLAM~\cite{matsuki2023gaussian}\!\!\!} & \multirow{3}{*}{\!\!\!\pub{CVPR24}\!} & \multirow{3}{*}{\checkmark} & PSNR$\uparrow$ & \textcolor{red}{34.83} & \textcolor{blue}{36.43} & \textcolor{blue}{37.49} & \textcolor{blue}{39.95} & \textcolor{blue}{42.09} & \textcolor{blue}{36.24} & \textcolor{red}{36.70} & \textcolor{blue}{36.07} & \textcolor{blue}{37.50} & \multirow{3}{*}{769} \\
         &  &  & SSIM$\uparrow$ & 0.95 & 0.96 & 0.96 & 0.97 & 0.98 & 0.96 & \textcolor{green}{0.96} & 0.96 & 0.96 & \\
         &  &  & LPIPS$\downarrow$ & \textcolor{red}{0.07} & \textcolor{green}{0.08} & \textcolor{red}{0.07} & \textcolor{green}{0.07} & \textcolor{blue}{0.06} & \textcolor{red}{0.08} & \textcolor{red}{0.07} & \textcolor{blue}{0.10} & \textcolor{red}{0.07} & \\
         \hline

         \multirow{3}{*}{Gaussian-SLAM~\cite{yugay2023gaussian}\!\!\!} & \multirow{3}{*}{\!\pub{arXiv}\!} & \multirow{3}{*}{\checkmark} & PSNR$\uparrow$ & \textcolor{blue}{34.31} & \textcolor{red}{37.28} & \textcolor{red}{38.18} & \textcolor{red}{43.97} & \textcolor{red}{43.56} & \textcolor{red}{37.39} & \textcolor{blue}{36.48} & \textcolor{red}{40.19} & \textcolor{red}{38.90} & \multirow{3}{*}{-} \\
         &  &  & SSIM$\uparrow$ & \textcolor{red}{0.99} & \textcolor{red}{0.99} & \textcolor{red}{0.99} & \textcolor{red}{1.00} & \textcolor{red}{0.99} & \textcolor{red}{0.99} & \textcolor{red}{0.99} & \textcolor{red}{1.00} & \textcolor{red}{0.99} & \\
         &  &  & LPIPS$\downarrow$ & \textcolor{green}{0.08} & \textcolor{red}{0.07} & \textcolor{red}{0.07} & \textcolor{red}{0.04} & \textcolor{green}{0.07} & \textcolor{red}{0.08} & \textcolor{blue}{0.08} & \textcolor{red}{0.07} & \textcolor{red}{0.07} & \\
         \hline
         
       \end{tabular}
    }
    \label{tbl:quanti_render_static}
\end{table*}


\begin{table*}[t]
    \begin{minipage}[t]{0.52\textwidth}
        \centering
        \captionsetup{width=0.98\linewidth}
        \caption{Comparison of \textbf{reconstruction} methods on D-NeRF~\cite{pumarola2021d} (dynamic scenes), in terms of PSNR, SSIM, and LPIPS. $^*$ denotes results reported in~\cite{wu20234d}.}
        \label{tbl:quanti_render_dynamic}
        \small
        \resizebox{\linewidth}{!}{ 
          \setlength\tabcolsep{6pt}{}
          \begin{tabular}{z{82}y{20}|c||ccc}
             \hline\thickhline
             \rowcolor{mygray}
             \multicolumn{2}{c|}{Method} & GS & PSNR$\uparrow$ & SSIM$\uparrow$ & LPIPS$\downarrow$ \\
             \hline\hline
             D-NeRF~\cite{pumarola2021d}\!\!&\!\!\!\pub{CVPR21}\! &  & 30.50 & 0.95 & 0.07 \\
         TiNeuVox-B~\cite{fang2022fast}\!\!&\!\!\!\pub{SGA22}\! &  & 32.67 & 0.97 & 0.04 \\
         KPlanes~\cite{fridovich2023k}\!\!&\!\!\!\pub{CVPR23}\! &  & 31.61 & 0.97 & - \\
         HexPlane-Slim~\cite{cao2023hexplane}\!\!&\!\!\!\pub{CVPR23}\! &  & 32.68 & 0.97 & 0.02 \\
         FFDNeRF~\cite{guo2023forward}\!\!&\!\!\!\pub{ICCV23}\! &  & 32.68 & 0.97 & 0.02 \\
         MSTH~\cite{wang2023masked}\!\!&\!\!\!\pub{NeurIPS23}\! &  & 31.34 & 0.98 & 0.02 \\
         3D GS$^*$~\cite{kerbl20233d}\!\!&\!\!\!\pub{TOG23}\! & \checkmark & 23.19 & 0.93 & 0.08 \\
         4DGS~\cite{yang2023real}\!\!&\!\!\!\pub{ICLR24}\! & \checkmark & 34.09 & 0.98 & - \\
         4D-GS~\cite{wu20234d}\!\!&\!\!\!\pub{CVPR24}\! & \checkmark & 34.05 & 0.98 & 0.02 \\
         GaGS~\cite{lu20243d}\!\!&\!\!\!\pub{CVPR24}\! & \checkmark & \textcolor{green}{37.36} & \textcolor{red}{0.99} & \textcolor{red}{0.01} \\
         CoGS~\cite{yu2023cogs}\!\!&\!\!\!\pub{CVPR24}\! & \checkmark & \textcolor{blue}{37.90} & 0.98 & 0.02 \\
         D-3DGS~\cite{yang2023deformable}\!\!&\!\!\!\pub{CVPR24}\! & \checkmark & \textcolor{red}{39.51} & \textcolor{red}{0.99} & \textcolor{red}{0.01} \\
         \hline
           \end{tabular}
        }
    \end{minipage}
    \hfill 
    \begin{minipage}[t]{0.45\textwidth}
        \centering
        \captionsetup{width=0.98\linewidth}
        \caption{Comparison of \textbf{reconstruction} methods on ZJU-MoCap~\cite{peng2021neural} (avatar), in terms of PSNR, SSIM, and LPIPS* (LPIPS $\times$ 1000). The results for non-GS methods are taken from \cite{lei2023gart}.}
        \label{tbl:quanti_avatar}
        \small
        \resizebox{\linewidth}{!}{ 
          \setlength\tabcolsep{4pt}{}
          \begin{tabular}{z{75}y{23}|c||ccc}
             \hline\thickhline
             \rowcolor{mygray}
             \multicolumn{2}{c|}{Method} & GS & PSNR$\uparrow$ & SSIM$\uparrow$ & LPIPS*$\downarrow$ \\
             \hline\hline
             NeuralBody~\cite{peng2021neural}\!\!\!&\!\pub{CVPR21}\! &  & 29.03 & 0.96 & 42.47 \\
         AnimNeRF~\cite{peng2021animatable}\!\!\!&\!\pub{ICCV21}\! &  & 29.77 & 0.96 & 46.89 \\
         PixelNeRF~\cite{yu2021pixelnerf}\!\!\!&\!\pub{ICCV21}\! &  & 24.71 & 0.89 & 121.86 \\
         NHP~\cite{kwon2021neural}\!\!\!&\!\pub{NeurIPS21}\! &  & 28.25 & 0.95 & 64.77 \\
         HumanNeRF~\cite{weng2022humannerf}\!\!\!&\!\pub{CVPR22}\! &  & 30.66 & 0.97 & 33.38 \\
         Instant-NVR~\cite{geng2023learning}\!\!\!&\!\pub{CVPR23}\! &  & \textcolor{green}{31.01} & 0.97 & 38.45 \\
         
         GauHuman~\cite{hu2023gauhuman}\!\!\!&\!\pub{CVPR24}\! & \checkmark & \textcolor{blue}{31.34} & 0.97 & \textcolor{green}{30.51} \\
         3DGS-Avatar~\cite{qian20233dgs}\!\!\!&\!\pub{CVPR24}\! & \checkmark & 30.61 & 0.97 & \textcolor{blue}{29.58} \\
         GART~\cite{lei2023gart}\!\!\!&\!\pub{CVPR24}\! & \checkmark & \textcolor{red}{32.22} & \textcolor{red}{0.98} & \textcolor{red}{29.21} \\
         \hline
           \end{tabular}
        }
    \end{minipage}
\end{table*}

\begin{wraptable}[11]{r}{0.52\textwidth}
    \centering
    \captionsetup{width=0.98\linewidth}
    \vspace{-6pt}
    \caption{Comparison of \textbf{reconstruction} methods on EndoNeRF~\cite{wang2022neural} (surgical scenes), in terms of PSNR, SSIM, and LPIPS. The results for non-GS methods are taken from~\cite{liu2024endogaussian}. FPS is measured using NVIDIA 4090~\cite{liu2024endogaussian}.}
    \vspace{-6pt}
    \small
    \resizebox{0.52\textwidth}{!}{
      \setlength\tabcolsep{3pt}{}
      \begin{tabular}{z{75}y{23}|c||ccc|cc}
         \hline\thickhline
         \rowcolor{mygray}
         \multicolumn{2}{c|}{Method} & GS & PSNR$\uparrow$ & SSIM$\uparrow$ & LPIPS$\downarrow$ & FPS$\uparrow$ & Mem.$\downarrow$ \\
         \hline\hline
         EndoNeRF~\cite{wang2022neural}\!\!\!&\!\pub{MICCAI22}\! &  & 36.06 & 0.93 & 0.09 & 0.04 & 19GB \\
         EndoSurf~\cite{zha2023endosurf}\!\!\!&\!\pub{MICCAI23}\! &  & 36.53 & 0.95 & 0.07 & 0.04 & 17GB \\
         LerPlane-9k~\cite{yang2023neural}\!\!\!&\!\pub{MICCAI23}\! &  & 35.00 & 0.93 & 0.08 & 0.91 & 20GB \\
         LerPlane-32k~\cite{yang2023neural}\!\!\!&\!\pub{MICCAI23}\! &  & \textcolor{green}{37.38} & 0.95 & 0.05 & 0.87 & 20GB \\

         Endo-4DGS~\cite{huang2024endo}\!\!\!&\!\pub{MICCAI24}\! & \checkmark & 37.00 & 0.96 & 0.05 & - & 4GB \\
         EndoGaussian~\cite{liu2024endogaussian}\!\!\!&\!\pub{arXiv}\! & \checkmark & \textcolor{blue}{37.85} & 0.96 & 0.05 & 195.09 & 2GB \\
         HFGS~\cite{zhao2024hfgs}\!\!\!&\!\pub{BMVC24}\! & \checkmark & \textcolor{red}{38.14} & \textcolor{red}{0.97} & \textcolor{red}{0.03} & - & - \\
         \hline
       \end{tabular}
    }
    \label{tbl:quanti_surgical}
\end{wraptable}

\end{document}